\documentclass[a4paper,11pt]{article}

\usepackage{graphicx}
\usepackage{epstopdf}
\usepackage{latexsym}
\usepackage{amsmath,amssymb}
\usepackage[ruled,vlined,linesnumbered]{algorithm2e}
\usepackage{color}
\usepackage{url}
\usepackage{hyperref}

\newcommand{\appref}[1]{Appendix~\ref{app:#1}}

\newcommand{\figref}[1]{Figure~\ref{fig:#1}}

\newcommand{\secref}[1]{Section~\ref{sec:#1}}
\newcommand{\tabref}[1]{Table~\ref{tab:#1}}

\newcommand{\bbC}{{\mathbb C}}
\newcommand{\bbZ}{{\mathbb Z}}
\newcommand{\bbR}{{\mathbb R}}

\newcommand{\MC}{{\mathcal C}}

\newcommand{\MF}{{\mathcal F}}
\newcommand{\MG}{{\mathcal G}}

\newcommand{\MT}{{\mathcal T}}

\newcommand{\ac}{\mathrm{ac}}
\newcommand{\bd}{\mathrm{bd}}
\newcommand{\cs}{\mathrm{cs}}
\newcommand{\dcp}{\mathrm{dcp}}
\newcommand{\dg}{\mathrm{dg}}
\newcommand{\ec}{\mathrm{ec}}
\newcommand{\ex}{\mathrm{ex}}
\newcommand{\fc}{\mathrm{fc}} 
\newcommand{\height}{\mathrm{ht}}
 
\newcommand{\inte}{\mathrm{int}}
\newcommand{\len}{\mathrm{len}}
\newcommand{\lf}{\mathrm{lf}}  
\newcommand{\mass}{\mathrm{mass}}
\newcommand{\na}{\mathrm{na}}

\newcommand{\rank}{\mathrm{rank}}
\newcommand{\val}{\mathrm{val}}
\newcommand{\rma}{\mathrm{a}}
\newcommand{\rmb}{\mathrm{b}}

\newcommand{\rmd}{\mathrm{d}}

\newcommand{\tta}{{\mathtt a}}
\newcommand{\ttb}{{\mathtt b}}
\newcommand{\ttC}{{\mathtt C}}
\newcommand{\ttCl}{\mathtt{Cl}}
\newcommand{\ttF}{{\mathtt F}}
\newcommand{\ttH}{{\mathtt H}}
\newcommand{\ttN}{{\mathtt N}}
\newcommand{\ttO}{{\mathtt O}}
\newcommand{\ttP}{{\mathtt P}}
\newcommand{\ttS}{{\mathtt S}}
\newcommand{\ttSi}{\mathtt{Si}}

\newcommand{\ul}[1]{\underline{#1}}
\newcommand{\Eex}{E^{\textrm{ex}}}
\newcommand{\Eint}{E^{\textrm{int}}}
\newcommand{\Eleaf}{E_{\textrm{leaf}}}
\newcommand{\fTL}{f_{\textrm{2L}}}
\newcommand{\fCC}{f_{\textrm{CC}}}
\newcommand{\Gac}{\Gamma_\mathrm{ac}} 
\newcommand{\hsup}[1]{\langle #1\rangle}
\newcommand{\Ldg}{\Lambda_{\mathrm{dg}}}
\newcommand{\molinf}{{\sf mol-infer}}
\newcommand{\oH}{\overline{{\tt H}}}  
\newcommand{\Vex}{V^{\textrm{ex}}}
\newcommand{\Vint}{V^{\textrm{int}}}
\newcommand{\Vleaf}{V_{\textrm{leaf}}}

\newcommand{\Cmin}{c_{\min}}
\newcommand{\Cmax}{c_{\max}}
\newcommand{\Vcirc}{V^{\circ}}
\newcommand{\Ecirc}{E^{\circ}}
\newcommand{\barEcirc}{\bar{E}^{\circ}}
\newcommand{\cc}{{\mathrm{cc}}}
\newcommand{\re}{{\rm edge}}
\newcommand{\rn}{{\rm node}}
\newcommand{\Mass}{\mathrm{Mass}}  
  
\newcommand{\ms}{\mathrm{ms}}  
  
\newcommand{\eledegfr}{\mathrm{eledeg}_\mathcal{F}}  
\newcommand{\ogamma}{{\overline{\gamma}}}
\newcommand{\otau}{{\overline{\tau}}}

\newcommand{\fr}{{\mathcal{F}}}
\newcommand{\msfr}{{\mathrm{ms}_\mathcal{F}}}
\newcommand{\htfr}{{\mathrm{ht}_\mathcal{F}}}
\newcommand{\valexfr}{\mathrm{val}_\mathcal{F}^{\mathrm{ex}}}  
\newcommand{\dfr}{\delta_\fr}
\newcommand{\aclf}{{\mathrm{ac}^{\mathrm{lf}}}}
\newcommand{\Gaclf}{{\Gamma_\mathrm{ac}^{\mathrm{lf}}}}
\newcommand{\Gacin}{{\Gamma_\mathrm{ac}^{\mathrm{int}}}}
\newcommand{\Gecin}{{\Gamma_\mathrm{ec}^{\mathrm{int}}}}
\newcommand{\aclfgamma}{{\mathrm{ac}_\gamma^{\mathrm{lf}}}}
\newcommand{\nint}{{n_\mathrm{int}}}
\newcommand{\nnoH}{{n_{\overline{\ttH}}}}
\newcommand{\nta}{{n_{\tta}}}
\newcommand{\ddg}{{\delta_{\mathrm{deg}}}}
\newcommand{\ddgint}{{\delta^{\mathrm{int}}_{\mathrm{deg}}}}
\newcommand{\dgnoH}{{\mathrm{deg}_{\overline{\ttH}}}}
\newcommand{\dgint}{{\mathrm{deg}^{\mathrm{int}}}}
\newcommand{\dbeta}{{\delta_\beta}}

\newcommand{\dac}{{\delta_{\mathrm ac}}}
\newcommand{\dec}{{\delta_{\mathrm ec}}}

\newcommand{\alphar}{{\alpha_{\mathrm{r}}}}
\newcommand{\dalpha}{{\delta_\alpha}}
\newcommand{\datom}{{\delta_{\mathrm{atm}}}}

\newcommand{\nain}{{\mathrm{na}^{\mathrm{int}}}}
\newcommand{\naex}{{\mathrm{na}^{\mathrm{ex}}}}
\newcommand{\nLB}{{n_{\mathrm{LB}}}}
\newcommand{\nUB}{{n_{\mathrm{UB}}}}
\newcommand{\naLB}{{\na_{\mathrm{LB}}}}
\newcommand{\naUB}{{\na_{\mathrm{UB}}}}
\newcommand{\ninLB}{{n^{\mathrm{int}}_{\mathrm{LB}}}}
\newcommand{\ninUB}{{n^{\mathrm{int}}_{\mathrm{UB}}}}
\newcommand{\nainLB}{{\na^{\mathrm{int}}_{\mathrm{LB}}}}
\newcommand{\nainUB}{{\na^{\mathrm{int}}_{\mathrm{UB}}}}
\newcommand{\naexLB}{{\na^{\mathrm{ex}}_{\mathrm{LB}}}}
\newcommand{\naexUB}{{\na^{\mathrm{ex}}_{\mathrm{UB}}}}
\newcommand{\acin}{{\mathrm{ac}^{\mathrm{int}}}}
\newcommand{\ecin}{{\mathrm{ec}^{\mathrm{int}}}}

\newcommand{\fcLB}{{\mathrm{fc}_{\mathrm{LB}}}}
\newcommand{\fcUB}{{\mathrm{fc}_{\mathrm{UB}}}}
\newcommand{\acLB}{{\mathrm{ac}^{\mathrm{int}}_{\mathrm{LB}}}}
\newcommand{\acUB}{{\mathrm{ac}^{\mathrm{int}}_{\mathrm{UB}}}}
\newcommand{\aclfLB}{{\mathrm{ac}^{\mathrm{lf}}_{\mathrm{LB}}}}
\newcommand{\aclfUB}{{\mathrm{ac}^{\mathrm{lf}}_{\mathrm{UB}}}}
\newcommand{\ecLB}{{\mathrm{ec}^{\mathrm{int}}_{\mathrm{LB}}}}
\newcommand{\ecUB}{{\mathrm{ec}^{\mathrm{int}}_{\mathrm{UB}}}}

\long\def\invis#1{}

\title{Cycle-Configuration: A Novel Graph-theoretic Descriptor Set for Molecular Inference\thanks{The work is partially supported by JSPS KAKENHI Grant Numbers JP22H00532 and JP22KJ1979.}}
\author{Bowen Song\and Jianshen Zhu\and Naveed Ahmed Azam\and
  Kazuya Haraguchi\and Liang Zhao\and Tatsuya Akutsu}
\date{}

\begin{document}
\maketitle

\begin{abstract}

In this paper, we propose a novel family of descriptors of chemical graphs,
named {\em cycle-configuration} ({\em CC}),
that can be used in the standard ``two-layered (2L) model'' of \molinf, a molecular inference framework
based on mixed integer linear programming (MILP)
and machine learning (ML).
Proposed descriptors capture the notion of
ortho/meta/para patterns that appear in aromatic rings,
which has been impossible in the framework so far. 
Computational experiments show that,
when the new descriptors are supplied,
we can construct prediction functions of similar or better
performance for all of the 27 tested chemical properties. 
We also provide an MILP formulation 
that asks for a chemical graph with desired properties
under the 2L model with CC descriptors (2L$+$CC model). 
We show that a chemical graph with up to 50 non-hydrogen vertices
can be inferred in a practical time. 
\end{abstract}

\section{Introduction}
\label{sec:intro}
Among key issues in cheminformatics and bioinformatics is
the problem of inferring molecules that are expected to
attain desired activities/properties.
This problem is also known as inverse QSAR/QSPR modeling~\cite{Ikebata17,Rupakheti15}.
We focus our attention on inverse QSAR/QSPR
modeling of low molecular weight organic compounds,
which has applications in drug discovery~\cite{Miyao16,Chembl.2023}\ 
and material science~\cite{Morgan.2020}. 
With recent rapid progress of machine learning (ML),
there have been developed a lot of inverse QSAR/QSPR models,
most of which are based on neural networks (NNs);
e.g., 
variational autoencoders~\cite{Gomez18},
generative adversarial networks~\cite{DeCao18, Prykhodko19},
and invertible flow models~\cite{Madhawa19, Shi20}.
The weakness of NN based methods is the lack of
optimality and exactness~\cite{ZhuJ:2023},
where we mean by optimality
the preciseness of a solution to attain the desired activities/properties;
and by exactness
the guarantee of a solution as a valid molecule. 
Besides, it is hard to exploit domain knowledge
in NN based methods. 

Our research group has developed a new framework
of molecular inference that is based on
mixed integer linear programming (MILP)
and ML.
This framework, which we call \molinf,
achieves optimality and exactness,
and enables practitioners to exploit domain knowledge
to some extent. 
Let $\MG$ denote the set of all possible chemical graphs. 
The process of \molinf\ is summarized as follows.
\begin{description}
\item[Stage 1:] Determine the target chemical property $\pi$
  and collect a data set $D_\pi\subseteq\MG$ of chemical graphs
  such that the observed value $a(\bbC)$ for the chemical property $\pi$
  is available for all chemical graphs $\bbC\in D_\pi$. 
\item[Stage 2:] Design a set of descriptors to obtain
  a feature function $f:\MG\to\bbR^K$
  that converts a chemical graph $\bbC\in D_\pi$ into
  a $K$-dimensional real feature vector $f(\bbC)\in\bbR^K$,
  where $K$ is the number of descriptors.
\item[Stage 3:] Construct a prediction function $\eta:\bbR^K\to\bbR$
  from the data set $f(D_\pi)\triangleq\{f(\bbC)\mid \bbC\in D_\pi\}$
  of feature vectors, where $\eta(\bbC)$
  is used to estimate the property value $a(\bbC)$
  of a chemical graph $\bbC$. 
\item[Stage 4:] Determine two real numbers $\underline{y}^\ast,\overline{y}^\ast$
  $(\underline{y}^\ast\le\overline{y}^\ast)$ as lower/upper bounds
  on the target value
  and 
  a set $\sigma$ of rules (called a {\em specification}) on chemical graphs.
  Let $\MG_\sigma\subseteq\MG$ denote the set of all chemical graphs
  that satisfy $\sigma$.
  Formulate the problem of constructing
  a chemical graph $\bbC^\dagger$ as MILP
  whose constraints include ${\mathcal C}_1$ and ${\mathcal C}_2$
  to ensure 
  $({\mathcal C}_1)$ $\underline{y}^\ast\le\eta(f(\bbC^\dagger))\le\overline{y}^\ast$
  and $({\mathcal C}_2)$ $\bbC^\dagger\in\MG_\sigma$. 
  Solve the MILP to obtain $\bbC^\dagger$. 
  If the MILP is infeasible,
  then it is indicated that no such $\bbC^\dagger$ exist. 
\item[Stage 5:] Generate isomers of $\bbC^\dagger$ somehow. 
\end{description}

Regarded as a method of inverse QSAR/QSPR,
the highlight of \molinf\ is Stage 4
that solves the inverse problem by MILP,
which is the original contribution of this framework.
For $\MC_1$, the process of computing
the feature vector $f(\bbC)$ for a chemical graph $\bbC$
and the process of computing
the prediction value $\eta(x)$ of a feature vector $x=f(\bbC)$
must be represented by 
linear inequalities of real and/or integer variables. 
It is shown that artificial neural network~\cite{AN.2019},
linear regression~\cite{LLR.2022} 
and decision tree~\cite{TZAHZNA21}
can be used as $\eta$.
We will discuss how to design $f$ for this purpose
in the next paragraph. 
For $\MC_2$, in our early studies, we could deal with only
limited classes of chemical graphs;
e.g., trees~\cite{ACZSNA20, Zhang.2022}, rank-1 graphs~\cite{Ito.2021}\ and rank-2 graphs~\cite{ZCSNA20}.
Shi et al.'s {\em two-layered} ({\em 2L}) {\em model}~\cite{SZAHZNA21}
admits us to infer {\em any} chemical graph,
where users are required to design an abstract structure of $\bbC^\dagger$
as a part of the specification $\sigma$. 
Stage 5 is not within the scope of this paper.
For this stage, a dynamic programming algorithm~\cite{Zhu.2021aa}\ %
and a grid neighborhood approach~\cite{gridAZHZNA21}
are developed.
Furthermore, \molinf\ is applied to the inference of polymers~\cite{poly_ICZAHZNA22}. 

Let us describe how we design the feature function $f$ in Stage 2.
The descriptors should be informative
since they have a great influence on
prediction performance in Stage 3
and thus on the quality of chemical graphs
that we finally obtain as a result of Stages 4 and 5;
if the prediction function $\eta$ is not accurate enough, then
we could not expect the inferred graphs to have desired property. 
On the other hand, as mentioned above,
the process of computing descriptor values should be
represented by a set of linear inequalities.
It is hard to include descriptors of complicated concepts.
There is a trade-off between informativity and simplicity
in the design of descriptors. 

\begin{figure}
  \centering
  \begin{tabular}{cc}
    \includegraphics[width=5cm]{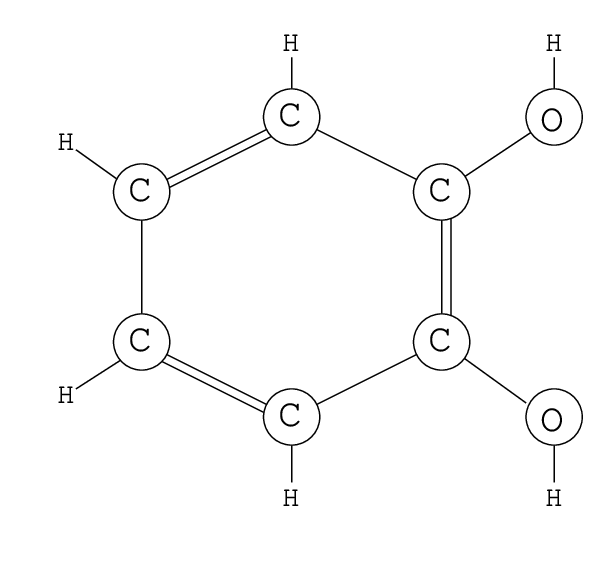} &
    \includegraphics[width=5cm]{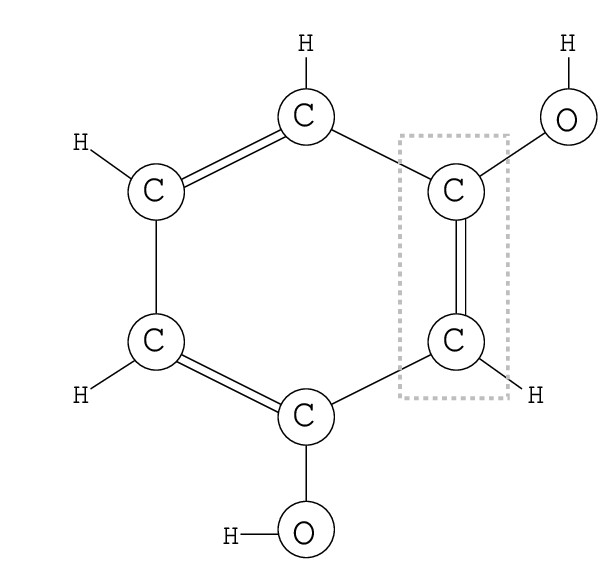} \\
                    {\bf (a)} & {\bf (b)} \\
                    \\
    \includegraphics[width=5cm]{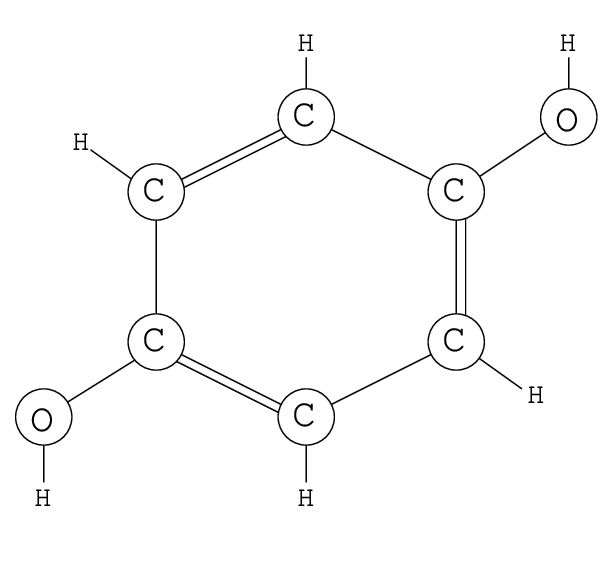} &
    \includegraphics[width=5cm]{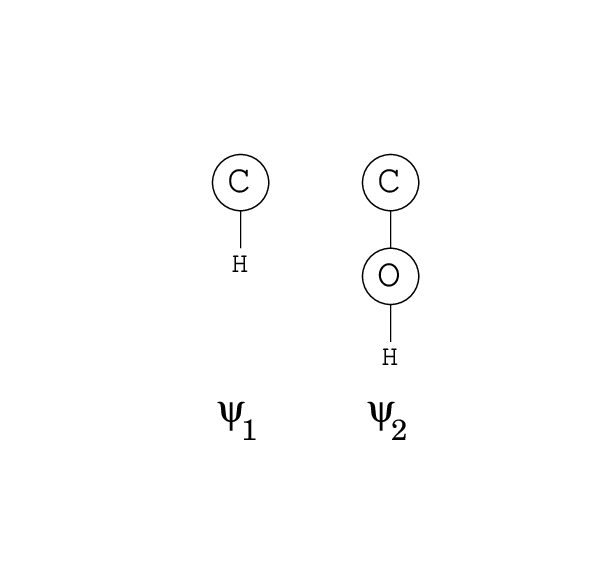}\\
                    {\bf (c)} & {\bf (d)}
  \end{tabular}
  \caption{(a) the chemical graph $\bbC_0$ for catechol;
    (b) the chemical graph $\bbC_1$ for resorcinol;
    (c) the chemical graph $\bbC_2$ for hydroquinone; and
    (d) two fringe-trees $\psi_1$ and $\psi_2$
    that appearing in all of $\bbC_0$, $\bbC_1$ and $\bbC_2$.
    In (b), the edge-configuration of the interior-edge indicated by a
    dotted rectangle is $(\ttC2,\ttC3,2)$.
    Although $\fTL(\bbC_1)=\fTL(\bbC_2)$,
    $a(\bbC_1)=0\ne1=a(\bbC_2)$ holds in
    the data set of AhR property from Tox21 collection.
  }
  \label{fig:weakpoints}
\end{figure}

Due to these reasons, 
\molinf\ employs graph-theoretic descriptors
that capture local information of chemical graphs
and that are somewhat similar to typical fingerprints. 
Let $\fTL$ be a feature function in the 2L model,
the standard model in \molinf.
The 2L model has a weak point such that
there are distinct chemical graphs $\bbC_1,\bbC_2$
for which $\fTL(\bbC_1)=\fTL(\bbC_2)$ holds
although $a(\bbC_1)$ and $a(\bbC_2)$ are much different.
An example of such $\bbC_1$ and $\bbC_2$
is shown in \figref{weakpoints},
where the details are explained in \secref{cc_motiv}. 
This issue comes from that the descriptors of the 2L model
cannot capture how edges are connected to cycles. 
For example, although the descriptors can distinguish ortho patterns
of an aromatic ring
(e.g., $\bbC_0$ in \figref{weakpoints})
from meta/para patterns (e.g., $\bbC_1$ and $\bbC_2$ in \figref{weakpoints}, respectively),
they fail to distinguish meta and para patterns.

In this paper, aiming at overcoming the above weak point in the 2L model,
we propose a novel set of descriptors, named {\em cycle-configurations}
({\em CC}).
CC can specify how exterior parts (called ``fringe-trees'')
are attached to a cycle,
by which meta/para patterns in an aromatic ring
are distinguishable. 
Let us denote by $\fCC$ a feature function that consists of CC descriptors.

We call the 2L model with CC descriptors the {\em 2L$+$CC model}.
In the 2L$+$CC model, 
we use the feature function $f:\MG\to\bbR^K$
such that $f(\bbC):=(\fTL(\bbC),\fCC(\bbC))$
for a chemical graph $\bbC$
(i.e., concatenation of two feature vectors $\fTL(\bbC)$ and $\fCC(\bbC)$).

Computational experiments show that, by using the 2L$+$CC model,
we can construct prediction functions of similar or better
performance for all of the 27 tested chemical properties,
in comparison with the 2L model.
We also provide an MILP formulation for the 2L$+$CC model
that asks for a chemical graph with desired properties. 
We show that a chemical graph with up to 50 non-hydrogen vertices
can be inferred in a couple of minutes.

The paper is organized as follows.
We make preparations and review the 2L model in \secref{prel}.
In \secref{cc}, we describe further background of CC
and provide its formal definition.
In \secref{milp}, we describe the idea of
the MILP for the 2L+CC model. 
We present computational results in \secref{exp}
and conclude the paper in \secref{conc}. 
Some details are explained in Appendix. 


\section{Preliminaries}
\label{sec:prel}

\subsection{Notations and Terminologies}
Let $\bbR$, $\bbR_+$, $\bbZ$ and $\bbZ_+$
denote the sets of reals,
non-negative reals, integers,
and non-negative integers, respectively.
For $p,q\in\bbZ$, let
us denote $[p,q]:=\{p,p+1,\dots,q\}$. 
For a vector (or a sequence) $x\in\bbR^p$ and $j\in[1,p]$,
we denote by $x(j)$ the $j$-th entry of $x$.
We denote $|x|:=p$. 

Let $A$ be a finite set.
To encode elements in $A$ by integers,
we may assume a bijection $\sigma:A\to[1,|A|]$ implicitly.
For $a\in A$, we represent the coded integer $\sigma(a)$
by $[a]_A$ or simply $[a]$
if $A$ is clear from the context.

For an undirected graph $G$, we denote by $V(G)$ and $E(G)$
the sets of vertices and edges, respectively.
For $V'\subseteq V(G)$
(resp., $E'\subseteq E(G)$),
we denote by $G-V'$ (resp., $G-E'$)
the subgraph of $G$ that is obtained by
removing the vertices in $V'$ along with
the incident edges (resp., removing the edges in $E'$).
When $V'=\{v\}$ (resp., $E'=\{e\}$),
we write $G-\{v\}$ as $G-v$ (resp., $G-\{e\}$ as $G-e$). 

A {\em cycle} $C$ in a graph $G$ is a subgraph of $G$ such that
$V(C)=\{u_1,u_2,\dots,u_{\ell}\}$ and $E(C)=\{u_1u_2,\dots,u_{\ell-1}u_{\ell},u_{\ell}u_1\}$.
We call $C$ {\em chordless} if there is no edge in $E(G)\setminus E(C)$
that joins vertices in $V(C)$. 
%
The length of a cycle $C$ is denoted by $\len(C)$
(i.e., $\len(C)=|V(C)|=|E(C)|=\ell$). 
When the length is $\ell$,
we call $C$ an {\em $\ell$-cycle}.

A graph is {\em rooted} if it has a designated vertex,
called a {\em root}.
For a graph $G$ possibly with a root,
a {\em leaf-vertex} is a non-root vertex with degree 1. 
We call the edge that is incident to a leaf-vertex 
a {\em leaf-edge}.
We denote by $\Vleaf(G)$
and $\Eleaf(G)$ the sets of
leaf-vertices and leaf-edges in $G$, respectively. 
For $i\in\bbZ_+$, we define the graph $G_i$
to be the subgraph of $G$ that is obtained by
deleting the set of leaf-vertices $i$ times,
that is, $G_0:=G$; and $G_{i+1}:=G_i-\Vleaf(G_i)$. 
We define the {\em height} $\height(v)$
of a vertex $v\in\Vleaf(G_i)$ to be $i$.
Note that the height is not defined for all vertices. 
\invis{
  We call a vertex $v$ a {\em tree vertex}
  if $v\in\Vleaf(G_i)$ for some $i\in\bbZ_+$.
  We define the {\em height} $\height(v)$
  of a tree vertex $v\in\Vleaf(G_i)$ to be $i$. 
  We do not define the height for non-tree vertices. 
}

\subsection{Modeling of Chemical Compounds}
We employ the modeling of chemical compounds that was introduced by
Zhu et al.~\cite{LLR.2022}. 

Let us represent chemical elements
by {\tt H} (hydrogen),
{\tt C} (carbon),
{\tt O} (oxygen),
{\tt N} (nitrogen) and so on.
To distinguish a chemical element
{\tt a} with multiple valences
such as {\tt S} (sulfur),
we denote {\tt a} with a valence $i$
by {\tt a}$_{(i)}$, where
we omit the suffix $(i)$ for a chemical
element with a unique valence. 
Let $\Lambda$ be a set
of  chemical elements;
e.g., $\Lambda=\{\ttH,\ttC,\ttO,\ttN,\ttP,\ttS_{(2)},\ttS_{(4)},\ttS_{(6)}\}$.
We represent the valence of ${\mathtt a}\in\Lambda$ by a function $\val:\Lambda\to[1,6]$; e.g.,
$\val(\ttH)=1$,
$\val(\ttC)=4$,
$\val(\ttO)=2$,
$\val(\ttP)=5$,
$\val(\ttS_{(2)})=2$ and
$\val(\ttS_{(6)})=6$. 
We denote the mass of $\tta\in\Lambda$ by $\mass^*(\tta)$. 

We represent a chemical compound by
a {\em chemical graph} that is defined to
be $\bbC=(H,\alpha,\beta)$ consisting of a simple, connected
undirected graph $H$ and functions
$\alpha:V(H)\to\Lambda$ and
$\beta:E(H)\to[1,3]$.
The set of atoms and the set of bonds
in the compound correspond to the vertex set $V(H)$
and the edge set $E(H)$, respectively.
The chemical element assigned to $v\in V(H)$
is represented by $\alpha(v)$
and the bond-multiplicity between two adjacent vertices
$u,v\in V(H)$ is represented by $\beta(e)$
of the edge $e=uv\in E(H)$.
We denote the mass of $H$ by $\mass^*(H):=\sum_{v\in V(H)}\mass^*(\alpha(v))$. 

Let $\bbC=(H,\alpha,\beta)$ be a chemical graph.
For a vertex $u\in V(H)$, we denote by $\beta_\bbC(u)$
the sum of bond-multiplicities of edges incident to $u$;
i.e., $\displaystyle\beta_\bbC(u):=\sum_{uv\in E(H)}\beta(uv)$.
We denote by $\deg_\bbC(u)$ the number of vertices
adjacent to $u$ in $\bbC$. 
For $\tta\in\Lambda$,
we denote by $V_{\tta}(\bbC)$ the set of
vertices in $v\in V(H)$ such that $\alpha(v)=\tta$ in $\bbC$.
We define the {\em hydrogen-suppressed chemical graph of $\bbC$},
denoted by $\hsup{\bbC}$, 
to be the graph that is obtained
by removing all vertices in $V_{\ttH}(\bbC)$ from $H$.

Two chemical graphs $\bbC_i=(H_i,\alpha_i,\beta_i)$,
$i=1,2$ are called {\em isomorphic}
if they admit an isomorphism,
i.e., a bijection $\phi:V(H_1)\to V(H_2)$ such that
``$uv\in E(H_1)$, $\alpha_1(u)=\tta$, $\alpha_1(v)=\ttb$,
$\beta_1(uv)=m$''
$\Leftrightarrow$
``$\phi(u)\phi(v)\in E(H_2)$,
$\alpha_2(\phi(u))=\tta$, $\alpha_2(\phi(v))=\ttb$,
$\beta_2(\phi(u)\phi(v))=m$''.
Furthermore,
when $H_i$ is a rooted graph
such that $r_i\in V(H_i)$ is the root, $i=1,2$,
$\bbC_1$ and $\bbC_2$ are called {\em rooted-isomorphic}
if they admit an isomorphism such that $\phi(r_1)=(r_2)$.

\subsection{Two-Layered (2L) Model}
We review the 2L model that was introduced by
Shi et al.~\cite{SZAHZNA21}. 

\subsubsection{Interior and Exterior}
Let $\bbC=(H,\alpha,\beta)$ be a chemical graph 
and $\rho\ge1$ be an integer, which we call a {\em branch-parameter},
where we use $\rho=2$ as the standard value. 
In the 2L model, the hydrogen-suppressed chemical graph
$\hsup{\bbC}$ is partitioned into ``interior'' and ``exterior'' parts
as follows. 
We call a vertex $v\in V(\hsup{\bbC})$
an {\em exterior-vertex} if $\height(v)<\rho$,
and an edge $e\in E(\hsup{\bbC})$
an {\em exterior-edge} if $e$ is incident to an exterior-vertex.
Let $\Vex(\bbC)$ and $\Eex(\bbC)$ denote the
sets of exterior-vertices and exterior-edges, respectively.
Define $\Vint(\bbC):=V(\hsup{\bbC})\setminus\Vex(\bbC)$ and
$\Eint(\bbC):=E(\hsup{\bbC})\setminus\Eex(\bbC)$.
We call a vertex in $\Vint(\bbC)$ an {\em interior-vertex}
and an edge in $\Eint(\bbC)$ an {\em interior-edge}.
We define the {\em interior} $\bbC^{\textrm{int}}$ {\em of $\bbC$}
to be the subgraph $(\Vint(\bbC),\Eint(\bbC))$. 

The set $\Eex(\bbC)$ of exterior-edges forms a collection
of connected graphs such that each is a tree $T$
rooted at an interior vertex $v\in V(T)$  
Let $\MT^{\textrm{ex}}(\hsup{\bbC})$ denote the family
of such chemical rooted trees in $\hsup{\bbC}$.
For each interior-vertex $u\in\Vint(\bbC)$,
let $T_u\in\MT^{\textrm{ex}}(\hsup{\bbC})$
denote the chemical tree rooted at $u$, where
$T_u$ may consist only of the vertex $u$.
We define the {\em fringe-tree of $u$},
denoted by $\bbC[u]$, to be the chemical rooted tree
that is obtained by putting back hydrogens to $T_u$
that are originally attached in $\bbC$. 

\subsubsection{Feature Function}
For a feature function $\fTL$ in the 2L model (Stage 2),
there are two types of descriptors:
static ones and enumerative ones.
There are 14 static descriptors such as
the number of non-hydrogen atoms and
the number of interior vertices.
The enumerative
descriptors mainly consist of the frequency of local patterns
that appear in a chemical graph $\bbC=(H,\alpha,\beta)$.
Examples of such local patterns
include ``fringe-configurations'',
``adjacency-configurations'' and
``edge-configurations''.
We collect enumerative descriptors from a given data set $D_\pi$. 

Let $u\in\Vint(\bbC)$ be an interior-vertex.
The {\em fringe-configuration of $u$}
is the chemical tree $\bbC[u]$ that is rooted at $u$.
Let us denote by $\MF(D_\pi)$ the set of all fringe trees
that appear in the data set $D_\pi$. 
For each $\psi\in\MF(D_\pi)$,
we introduce a descriptor that evaluates the number of interior-vertices
$u\in\Vint(\bbC)$ such that $\bbC[u]$ is rooted-isomorphic to $\psi$.

For an interior-edge $e=uv\in\Eint(\bbC)$,
let $\alpha(u)=\tta$, $\deg_{\hsup{\bbC}}(u)=d$,
$\alpha(v)=\ttb$, $\deg_{\hsup{\bbC}}(v)=d'$ and $\beta(e)=m$.
The {\em adjacency-configuration of $e$} (resp., {\em edge-configuration of $e$})
is defined to be the tuple $(\tta,\ttb,m)$ (resp., $(\tta d,\ttb d',m)$). 
Let us denote by $\Gamma_{\textrm{ac}}(D_\pi)$ (resp., $\Gamma_{\textrm{ec}}(D_\pi)$)
the set of all adjacency-configurations (resp., edge-configurations)
in the data set $D_\pi$. 
For each tuple $\gamma_{\textrm{ac}}\in\Gamma_{\textrm{ac}}(D_\pi)$
(resp., $\gamma_{\textrm{ec}}\in\Gamma_{\textrm{ec}}(D_\pi)$),
we introduce a descriptor that evaluates the number of interior-edges
$e\in\Eint(\bbC)$ such that the adjacency-configuration (resp., edge-configuration)
is equal to $\gamma_{\textrm{ac}}$ (resp., $\gamma_{\textrm{ec}}$). 

See \appref{2Lmodel} 
for a full description of descriptors in the 2L model.

\subsubsection{Specification for MILP}
\label{sec:prel_MILP}
In the 2L model,
the specification $\sigma$ for MILP (Stage 4)
consists of the following three rules: 
\begin{itemize}
\item a {\em seed graph} $G_{\textrm{C}}$ as an abstract form of a
  target chemical graph $\bbC^\dagger$; 
\item a set $\MF$ of fringe trees
  as candidates for a tree $\bbC^\dagger[u]$ rooted
  at each interior-vertex in $\bbC^\dagger$; and
\item lower/upper bounds on
  the number of various parameters in $\bbC^\dagger$;
  e.g., chemical elements, double/triple bonds, and 
  fringe/edge/adjacency-configurations. 
\end{itemize}
The MILP formulates the process of constructing a chemical graph $\bbC^\dagger$
as follows. First, we 
decide the interior of $\bbC^\dagger$ by ``expanding'' the seed graph $G_{\textrm{C}}$;
e.g., subdividing an edge and
attaching a new path to a vertex. 
Second, regarding all vertices in the expanded seed graph
as the interior-vertices of $\bbC^\dagger$, 
we assign a fringe tree in $\MF$ to every vertex
to make the exterior of $\bbC^\dagger$. 
Finally, we assign 
bond-multiplicities to the interior-edges
so that all constraints in $\sigma$ are satisfied.
We can regard $\MG_\sigma$ in \secref{intro}
as the set of all chemical graphs that can be
constructed in this way. 
See the preprint of \cite{LLR.2022}
for details of MILP in the 2L model. 



\section{Cycle-Configurations}
\label{sec:cc}
In this section, we propose a new type of descriptors
for the 2L model, named cycle-configurations (CC).

\subsection{Motivation}
\label{sec:cc_motiv}
Let us point out a weak point of the 2L model again;
there are chemical graphs that are not isomorphic to each other
but are converted into an identical feature vector. 
See \figref{weakpoints} for an example.
Three chemical graphs $\bbC_0$ (catechol), $\bbC_1$
(resorcinol) and $\bbC_2$ (hydroquinone) are shown,
where $\bbC_0$ is the ortho-isomer, 
$\bbC_1$ is the meta-isomer and $\bbC_2$ is the para-isomer. 

We can confirm that $\fTL(\bbC_1)=\fTL(\bbC_2)$ holds
by observing the descriptors one by one. 
For fringe-configuration,
both chemical graphs contain
four $\psi_1$ and two $\psi_2$ as fringe-trees in common. 
For edge-configuration,
they contain
one $(\ttC2,\ttC2,1)$;
one $(\ttC2,\ttC2,2)$;
two $(\ttC2,\ttC3,1)$ and
two $(\ttC2,\ttC3,2)$ in common. 
In this way, one sees that the two chemical graphs
take the same values for the other descriptors (see \appref{2Lmodel}). 
We also see that $\fTL(\bbC_0)\ne\fTL(\bbC_1)$
and $\fTL(\bbC_0)\ne\fTL(\bbC_2)$ hold
since $\bbC_0$ contains
one $(\ttC2,\ttC2,1)$;
two $(\ttC2,\ttC2,2)$;
two $(\ttC2,\ttC3,1)$ and
one $(\ttC3,\ttC3,2)$
for its edge-configurations, which are different from those of $\bbC_1$ and $\bbC_2$. 

Although the two chemical graphs $\bbC_1$ and $\bbC_2$
are converted into an identical feature vector,
they may have different properties from each other.
For example, Tox21 is a collection of
data sets for binary classification
(i.e., $a(\bbC)\in\{0,1\}$ for $\bbC\in D_\pi$).
In AhR data set, 
the two chemical graphs $\bbC_1$ and $\bbC_2$ in \figref{weakpoints}
satisfy $\fTL(\bbC_1)=\fTL(\bbC_2)$ although $a(\bbC_1)=0$ and $a(\bbC_2)=1$ hold.
It is desirable to convert as many such pairs
into distinct feature vectors as possible.

\subsection{Definitions}
We define a new descriptor,
cycle-configuration, in order to convert chemical graphs like $\bbC_1$
and $\bbC_2$ in \figref{weakpoints} into distinct feature vectors. 
Let $R=\{a_1,a_2,\dots,a_k\}$ be a set of distinct real numbers. 
For $a\in R$, we define $\rank_R(a):=i$ if 
$a$ is the $i$-th smallest in $R$. 
For example, when $R=\{3,2,5,9\}$,
we have $\rank_R(3)=2$, $\rank_R(2)=1$,
$\rank_R(5)=3$, and $\rank_R(9)=4$.

Suppose that a chemical graph $\bbC$ is given. 
Let $C$ be a chordless cycle in $\bbC$
such that $V(C)=\{u_1,u_2,\dots,u_{\ell}\}$ and 
$E(C)=\{u_1u_2,u_2u_3,\dots,u_{\ell-1}u_{\ell},u_{\ell}u_1\}$.  
For $u_i\in V(C)$, we define $\mu_i:=\mass^\ast(\bbC[u_i])$. 
Let $R$ denote the set of distinct numbers in $\mu_1,\mu_2,\dots,\mu_\ell$.
We define $\xi(C)$ to be the smallest sequence
$(\rank_R(\mu_1),\rank_R(\mu_2),\dots,\rank_R(\mu_{\ell}))$
with respect to the lexicographic order
among all possible cyclic permutations (including reversal)
of $(u_1,u_2,\dots,u_{\ell})$, 
where there are $2\ell$ permutations possible. 
We define the {\em cycle-configuration of $C$} to be $\xi(C)$.

Let us see \figref{weakpoints} for example. 
Suppose $\mass^\ast(\ttH)=1$, $\mass^\ast(\ttC)=12$ and $\mass^\ast(\ttO)=16$.
For the two fringe-trees $\psi_1$ and $\psi_2$ in the figure, 
we have $\mass^\ast(\psi_1)=12+1=13$ and $\mass^\ast(\psi_2)=12+16+1=29$. 
Let us denote the unique (chordless) 6-cycle in $\bbC_i$ by $C_i$, $i=1,2$. 
The set of distinct numbers that appear as the mass of a fringe-tree 
is $R=\{13,29\}$ for both chordless cycles,
where $\rank_R(13)=1$ and $\rank_R(29)=2$. 
One readily sees that $\xi(C_1)=\xi_1:=(1,1,1,2,1,2)$ and $\xi(C_2)=\xi_2:=(1,1,2,1,1,2)$. 

CCs are enumerative descriptors,
and we collect ones that are included in the feature function
from a given data set $D_\pi$.
We denote by $\Xi(D_\pi)$
the set of all cycle-configurations $\xi$ that appear in $D_\pi$. 
Let $K_{\textrm{CC}}:=|\Xi(D_\pi)|$.
For a chemical graph $\bbC\in D_\pi$,
we define $f_{\textrm{CC}}(\bbC)$ to be
a $K_{\textrm{CC}}$-dimensional feature vector
$f_{\textrm{CC}}(\bbC)=(\dcp^\circ_1(\bbC),\dcp^\circ_2(\bbC),\dots,\dcp^\circ_{K_{\textrm{CC}}}(\bbC))$,
where
\begin{quote}
  $\dcp^\circ_i(\bbC)$, $i=[\xi^\ast]$, $\xi^\ast\in\Xi(D_\pi)$:
  the number of chordless cycles $C$ in $\bbC$
  such that $\xi(C)=\xi^\ast$. 
\end{quote}

See \figref{weakpoints} again. 
Suppose $\Xi(D_\pi)=\{\xi_1,\xi_2,\xi_3,\xi_4\}$
for $\xi_3:=(1,1,2,3)$ and $\xi_4:=(1,1,1,1,2)$.
Then $\fCC(\bbC_1)=(1,0,0,0)$ and $\fCC(\bbC_2)=(0,1,0,0)$ hold,
by which we have $f(\bbC_1)=(\fTL(\bbC_1),\fCC(\bbC_1))\ne (\fTL(\bbC_2),\fCC(\bbC_2))=f(\bbC_2)$. 

In our implementation,
as $D_\pi$ may contain too many CC descriptors,
we use only CC descriptors whose lengths are in the range $[c_{\min},c_{\max}]$,
where $c_{\min}$ and $c_{\max}$ are positive constants $(c_{\min}\le c_{\max})$.
We will set $c_{\min}:=4$ and $c_{\max}:=6$
since, in most of chemical compounds
in conventional databases, 
the chemical graph
is acyclic or contain only chordless cycles
whose lengths are within $[4,6]$. 
See \tabref{database}.
For example, 
in PubChem,
among 92,509,596 molecules that are feasible in the 2L-model,
83,520,760 molecules (90\%) satisfy this condition. 
    
\begin{table}[t]
  \centering
  \caption{The numbers of chemical compounds in conventional databases. The 2nd to 5th columns represent the number of all registered chemical compounds; the number of feasible chemical graphs in the 2L-model (e.g., connected, at least four carbon atoms exist); the number of chemical graphs that are either acyclic or $\ell(C)\in[4,6]$ for all chordless cycles $C$; the number of chemical graphs that contain none of (i) or (ii), respectively. The percentages indicate the ratio of the number over the left number. }
  \label{tab:database}
  {\small
  \begin{tabular}{lrrrr}
    \hline
    Database &\multicolumn{1}{c}{All}
    &\multicolumn{1}{c}{2L-model}
    &\multicolumn{1}{c}{Acyclic or $\ell(C)\in[4,6]$}
    &\multicolumn{1}{c}{No substructures}\\
    &\multicolumn{1}{c}{}
    &\multicolumn{1}{c}{feasible}
    &\multicolumn{1}{c}{for all chordless cycles $C$}
    &\multicolumn{1}{c}{(i) or (ii) in \secref{milp}}\\
    \hline
    PubChem & 97,092,888 & 92,509,596 & 83,520,760 & 80,842,345 \\
    (as of 2019)&& (95\%) & (90\%) & (96\%) \\
    QM9 & 130,786 & 130,786 & 71,520 & 60,352 \\
    && (100\%) & (54\%) & (84\%)  \\
    Tox21 & 8,014 & 7,769 & 7,273 & 7,080 \\
    && (96\%) & (93\%) & (97\%)   \\
    \hline
  \end{tabular}
  }
\end{table}


\section{MILP Formulation for 2L$+$CC Model}
\label{sec:milp}

Let us consider an MILP formulation
for inferring a chemical graph in the 2L$+$CC model.
Similarly to the 2L model,
the constraints of the MILP consist of $(\MC_1)$
$\underline{y}^\ast\le\eta(f(\bbC^\dagger))\le\overline{y}^\ast$
and $(\MC_2)$ $\bbC^\dagger\in\MG_\sigma$,
where $\bbC^\dagger$ denotes a chemical graph to be inferred
and is represented by real/integer variables. 
We can use any prediction function $\eta$ in $\MC_1$
if its computational process can be represented by
a set of linear inequalities. For example,
artificial neural network~\cite{AN.2019},
linear regression~\cite{LLR.2022} 
and decision tree~\cite{TZAHZNA21} can be used
to construct $\eta$.  
In this section, we overview how we formulate $\MC_2$ as MILP. 
See \appref{MILP} for the precise formulation of the MILP
that includes how we represent 
the computational process of the
feature function $f$ by
a set of linear inequalities.

The basic idea of $\MC_2$ is 
similar to the 2L model (see \secref{prel_MILP});
we represent by $\MC_2$
the computational process of 
expanding an abstract form of the chemical graph 
to a concrete chemical graph.
We introduce a new type of abstract form,
which we call a ``seed tree'',
since it is hard to deal with CC descriptors
by a seed graph of the 2L model.

\begin{figure}[t!]
  \centering
  \begin{tabular}{ccc}
    \includegraphics[width=3.5cm]{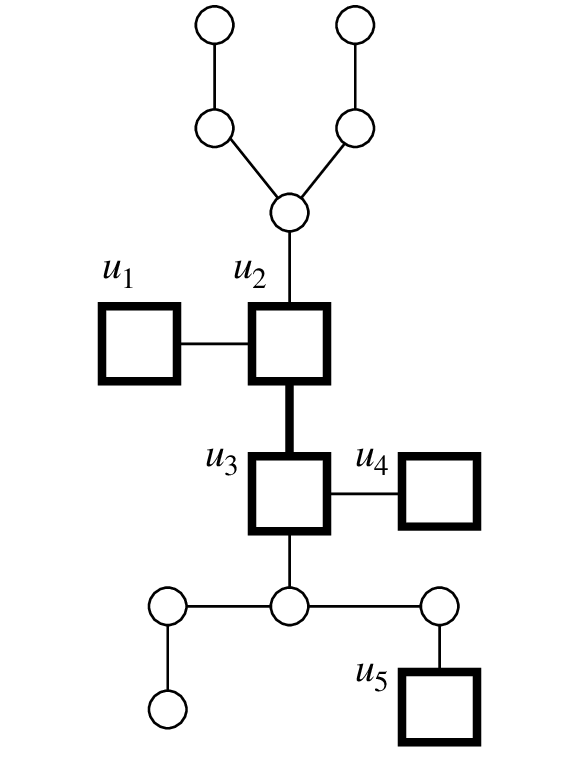} &
    \includegraphics[width=3.5cm]{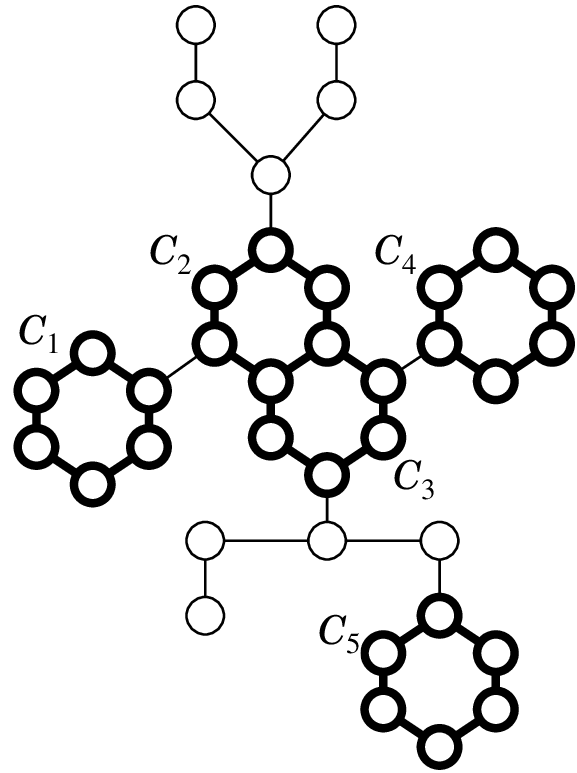} &
    \includegraphics[width=7cm]{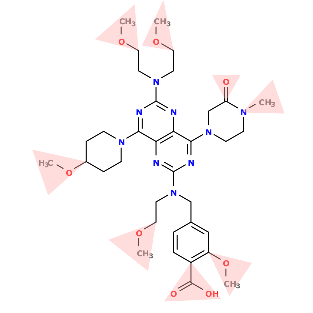}\\
                    {\bf (a)} & {\bf (b)} & {\bf (c)}
  \end{tabular}
  \caption{Construction of a chemical graph. (a) A seed tree. Thick squares/lines indicate ring nodes/edges, while thin circles/lines indicate non-ring nodes/edges.
    (b) Ring nodes are expanded to chordless 6-cycles.
    (c) Fringe-trees are assigned to every vertex and bond-multiplicities are assigned to every edge.
    Fringe-trees of non-zero heights are indicated by shade. 
    The PubChem CID of the compound is 156839899, and the molecular formula is C$_{35}$H$_{51}$N$_9$O$_8$.}
  \label{fig:seed_tree}
\end{figure}

A {\em seed tree} is a tuple $\MT=(T;V^{\circ},E^{\circ})$
of an unrooted tree $T$, 
$\Vcirc\subseteq V(T)$ and $\Ecirc\subseteq \{uv\in E(T)\mid u,v\in\Vcirc\}$.
We call a node in $\Vcirc$ a {\em ring node} and
an edge in $\Ecirc$ a {\em ring edge},
whereas a node in $V(T)\setminus \Vcirc$ is a {\em non-ring node},
and an edge in $E(T)\setminus \Ecirc$ is a {\em non-ring edge}. 
For a node $u\in V(T)$,
we denote by $\Ecirc(u)$ and $\barEcirc(u)$
the sets of all ring edges
and of all non-ring edges incident to $u$, respectively.
See \figref{seed_tree}(a) for an example. 

We formulate by $\MC_2$
the following process of constructing
a chemical graph $\bbC^\dagger:=\bbC_{\MT}=(G_{\MT},\alpha,\beta)$.

\begin{description}
\item[(I)] Each ring node $u\in\Vcirc$
  is assigned a cycle-configuration,
  by which $u$ is ``expanded'' to a chordless cycle in $G_{\MT}$. 
  \begin{itemize}
  \item If two ring nodes are joined by a ring edge,
    then the corresponding two chordless cycles in $G_{\MT}$ share an edge in common.
  \item Each non-ring node in $V(T)\setminus \Vcirc$
    appears as a single vertex in $G_{\MT}$.
  \item Each non-ring edge in $E(T)\setminus\Ecirc$
    appears as a single edge in $G_{\MT}$.
  \end{itemize}
\item[(II)] The expanded graph is used as the interior of $G_{\MT}$.
  For the exterior, fringe-trees are assigned to all nodes in the expanded graph,
  and to the interior-edges, bond-multiplicities are assigned. 
\end{description}


For the seed tree $\MT=(T;\Vcirc,\Ecirc)$
in \figref{seed_tree}(a), we have $\Vcirc=\{u_1,u_2,\dots,u_5\}$. 
Suppose that we are given $\Xi^u=\{\xi_1,\xi_2,\dots,\xi_6\}$
for all ring nodes $u\in \Vcirc$, where
\begin{align*}
  &\xi_1=(1,1,2,3), &&\xi_2=(1,1,1,1,2), &&\xi_3=(1,1,1,1,1,2),\\
  &\xi_4=(1,1,1,1,2,3), &&\xi_5=(1,1,1,2,1,2), &&\xi_6=(1,2,2,4,3,2).
\end{align*}
In this example, $u_1$ is assigned $\xi_3$;
$u_2$ and $u_3$ are assigned $\xi_5$;
$u_4$ is assigned $\xi_4$; and
$u_5$ is assigned $\xi_6$, where $\xi_1$ and $\xi_2$ are assigned to no ring nodes. 
As shown in \figref{seed_tree}(b),
all ring nodes are expanded to chordless 6-cycles, $C_1$ to $C_5$,
where $6=|\xi_3|=|\xi_4|=|\xi_5|=|\xi_6|$.
We can confirm that the CCs of the five corresponding
chordless cycles in \figref{seed_tree}(c)
are precisely ones that are assigned above.
For example, in $C_4$,
there are four distinct fringe-trees 
whose molecular formulas are N, CH$_2$, CO, CH$_3$N,
where we denote them by $\psi_1,\psi_2,\psi_3,\psi_4$, respectively.
We have $\mass^\ast(\psi_1)=14$, $\mass^\ast(\psi_2)=12+2=14$, $\mass^\ast(\psi_3)=12+16=28$
and $\mass^\ast(\psi_4)=12+3+14=29$,
where $\mass^\ast(\psi_1)=\mass^\ast(\psi_2)=14$,
and hence $\xi(C_4)=(1,1,1,1,2,3)=\xi_4$ holds. 

As we observed in \secref{cc},
CC descriptors can distinct how exteriors are attached to
a chordless cycle in a chemical graph (e.g., meta/para-isomers of an aromatic ring),
which is impossible by the original descriptors in the 2L model. 
Ring nodes are, however, not necessarily universal;
there is a set of chordless cycles
that cannot be represented by expanding ring nodes. 
For example:
%
\begin{description}
\item[(i)] A pair of two chordless cycles that share exactly one point $v$. 
\item[(ii)] A set of more than two chordless cycles 
  that share one vertex or edge in common. 
\end{description}
To infer $\bbC^\dagger$ that contains at least one of the above structures, 
one needs to make use of non-ring nodes/edges appropriately
in the design of a seed tree.
We note that, however,
such chemical compounds are rather minor
in conventional databases.
See \tabref{database} again.
For example, in PubChem,
among 83,520,760 molecules that are acyclic
or contain only chordless cycles whose lengths are within $[4,6]$, 
80,842,345 molecules (96\%) contain neither (i) nor (ii).

\tabref{spec} shows a description of the specification $\sigma$ in the 2L$+$CC model.
Besides the seed tree, 
the specification
includes availability of chemical elements/configurations,
lower/upper bounds on their numbers. 

\begin{table}[t!]
  \centering
  \caption{A description of specification $\sigma$ in the 2L$+$CC model (AC: adjacency-configuration; CC: cycle-configuration; EC: edge-configuration; FC: fringe-configuration)}
  \label{tab:spec}
  \begin{tabular}{ll}
    \hline
        {\bf Symbol}&
        {\bf Definition}\\
        \hline
        $\MT=(T;V^{\circ},E^{\circ})$ & A seed tree \\
        \hline
        \multicolumn{2}{l}{\bf (A set of available chemical elements/configurations in $G_{\MT}$)}\\
        $\Lambda$ & Chemical elements\\
        $\Xi^u$ & CCs for $u\in\Vcirc$, where $\xi\in\Xi^u$ satisfies $\Cmin\le|\xi|\le\Cmax$\\
        $\fr^u$ & FCs for $u\in V(T)$\\
        $\Gacin$ & ACs on interior-edges\\
        $\Gaclf$ & ACs on leaf-edges\\
        $\Gecin$ & ECs on interior-edges\\
        \hline
        \multicolumn{2}{l}{\bf (Lower/upper bounds on the numbers in $G_{\MT}$)}\\
        $\nLB, \nUB$& The number of non-hydrogen atoms\\
        $\nainLB(\tta), \nainUB(\tta)$ & The number of chemical elements $\tta\in\Lambda$ in the interior\\ 
        $\naexLB(\tta), \naexUB(\tta)$ & The number of chemical elements $\tta\in\Lambda$ in the exterior\\ 
        $\naLB(\tta), \naUB(\tta)$ & The number of chemical elements $\tta\in\Lambda$ in $G_\MT$\\ 
        $\fcLB(\psi), \fcUB(\psi)$ & The number of FCs $\psi \in \fr^\MT:=\bigcup_{u\in V(T)}\fr^u$\\       
        $\acLB(\gamma), \acUB(\gamma)$ & The number of ACs $\gamma\in\Gacin$ in interior-edges\\ 
        $\aclfLB(\gamma), \aclfUB(\gamma)$ & The number of ACs $\gamma\in\Gaclf$
        in leaf-edges\\ 
        $\ecLB(\gamma), \ecUB(\gamma)$ & The number of ECs $\gamma\in\Gecin$ in interior-edges\\ 
        \hline
  \end{tabular}
\end{table}


\section{Computational Experiments}
\label{sec:exp}

In this section, we describe experimental results on
Stages 3 (ML) and 4 (MILP) in \molinf. 
All experiments are conducted on a PC
that carries Apple Silicon M1 CPU
(3.2GHz)
and 8GB main memory. 
All source codes are written in Python
with a machine learning library {\tt scikit-learn} of version 1.5.0.
The source codes and results are available
at \url{https://github.com/ku-dml/mol-infer/tree/master/2LCC}. 

\subsection{Experimental Setup (Stages 1 and 2)}
We collected data sets for 27 chemical properties
that are shown in \tabref{data_reg}. 
In these data sets,
the property value $a(\bbC)$ of a chemical graph $\bbC$
is a real number, and hence the ML task in Stage 3 is regression. 
QM9 properties taken from \cite{molnet}
(i.e., {\sc Alpha}, {\sc Cv}, {\sc Gap}, {\sc Homo},
{\sc Lumo}, {\sc mu} and {\sc U0}) share the same data set
in common. This original data set contains more than $1.3\times10^5$ molecules
and we use a subset of $10^3$ molecules that are randomly selected. 

From the original data set,
we exclude molecules that are not feasible in the 2L model
(e.g., the chemical graph is not connected). 
Furthermore,
we decide the set $\Lambda$ of available chemical elements
for each property $\pi$,
by which chemical graphs that contain rare chemical elements are eliminated.

Details of columns in \tabref{data_reg}
are described as follows. 
\begin{itemize}
\item $\Lambda\setminus\{\ttH\}$:
  the set of available chemical elements except hydrogen,
  where \\
  $\Lambda_1=\{\ttC_{(2)},\ttC_{(3)},\ttC_{(4)},\ttC_{(5)},\ttO,\ttN_{(1)},\ttN_{(2)},\ttN_{(3)},\ttF\}$;
  $\Lambda_2=\{\ttC,\ttO,\ttN,\ttS_{(2)},\ttS_{(6)},\ttCl\}$;
  $\Lambda_3=\{\ttC,\ttO\}$;
  $\Lambda_4=\{\ttC,\ttO,\ttN\}$;
  $\Lambda_5=\{\ttC_{(2)},\ttC_{(3)},\ttC_{(4)},\ttO,\ttN_{(2)},\ttN_{(3)},\ttS_{(2)},\ttS_{(4)}\ttS_{(6)},\ttCl\}$;
  $\Lambda_6=\{\ttC,\ttO,\ttN,\ttS_{(2)},$ $\ttS_{(4)},\ttS_{(6)},\ttCl\}$; and 
  $\Lambda_7=\{\ttC,\ttO,\ttSi\}$.  
\item $\underline{n}$ and $\overline{n}$:
  the minimum and maximum values
  of the number of non-hydrogen atoms in $\bbC\in D_\pi$.
\item $|D_\pi|$: the number of chemical graphs in the data set.
\item $K_{\textrm{2L}}$ and $K_{\textrm{CC}}$:
  the number of 2L and CC descriptors extracted from $D_\pi$, respectively. 
\end{itemize}

\begin{table}[t]
  \centering
  \caption{Summary of data sets}
  \label{tab:data_reg}
  {\scriptsize
  \begin{tabular}{lcccrrr}
    \hline
    $\pi$ (Description) & Ref. & $\Lambda\setminus\{\ttH\}$
    & $\underline{n},\overline{n}$
    & \multicolumn{1}{c}{$|D_\pi|$}
    & \multicolumn{1}{c}{$K_{\textrm{2L}}$}
    & \multicolumn{1}{c}{$K_{\textrm{CC}}$}\\
    \hline
    {\sc Alpha} (Isotropic polarizability)
    & \cite{molnet} & $\Lambda_1$ & 6,9& 977 & 297 & 184 \\
    {\sc At} (Autoignition temperature)
    & \cite{pubchem} & $\Lambda_2$ & 4,85 & 448 & 255 & 65 \\
    {\sc Bhl} (Biological half life)
    & \cite{pubchem} & $\Lambda_2$ & 5,36 & 514 & 166 & 94 \\
    {\sc Bp} (Boiling point)
    & \cite{pubchem} & $\Lambda_2$ & 4,67 & 444 & 230 & 70 \\
    {\sc Cv} (Heat Capacity at 298.15K)
    & \cite{molnet} & $\Lambda_1$ & 6,9 & 977 & 297 & 184 \\
    {\sc Dc} (Dissociation constants)
    & \cite{pubchem} & $\Lambda_2$ & 5,44 & 161 & 130 & 63 \\
    {\sc EDPA} (Electron density on the most positive atom)
    & \cite{KOVATS.2001} & $\Lambda_3$ & 11,16 & 52 & 64 & 6 \\
    {\sc Fp} (Flash point in closed cup)
    & \cite{pubchem} & $\Lambda_2$ & 4,67& 424 & 229 & 70 \\
    {\sc Gap} (Gap between {\sc Homo} and {\sc Lumo})
    & \cite{molnet} & $\Lambda_1$ & 6,9 & 977 & 297 & 184 \\
    {\sc Hc} (Heat of combustion)
    & \cite{pubchem} & $\Lambda_2$ & 4,63 & 282 & 177 & 49 \\
    {\sc Homo} (Energy of highest occupied molecular orbital)
    & \cite{molnet} & $\Lambda_1$ & 6,9 & 977 & 297 & 184\\
    {\sc Hv} (Heat of vaporization)
    & \cite{pubchem} & $\Lambda_4$ & 4,16 & 95 & 105 & 16 \\
    {\sc IhcLiq} (Isobaric heat capacities; liquid)
    & \cite{N.2019} & $\Lambda_4$ & 4,78 & 770 & 256 & 74 \\
    {\sc IhcSol} (Isobaric heat capacities; solid)
    & \cite{N.2019} & $\Lambda_5$ & 5,70 & 668 & 228 & 118 \\
    {\sc KovRI} (Kovats retention index)
    & \cite{KOVATS.2001} & $\Lambda_3$ & 11,16 & 52 & 64 & 6\\
    {\sc Kow} (Octanol/water partition coefficient)
    & \cite{pubchem} & $\Lambda_4$ & 4,58 & 684 & 223 & 117 \\
    {\sc Lp} (Lipophilicity)
    & \cite{molnet} & $\Lambda_6$ & 6,74 & 936 & 231 & 178 \\
    {\sc Lumo} (Energy of lowest occupied molecular orbital)
    & \cite{molnet} & $\Lambda_1$ & 6,9 & 977 & 297 & 184 \\
    {\sc Mp} (Melting point)
    & \cite{pubchem} & $\Lambda_6$ & 4,122 & 577 & 255 & 108 \\
    {\sc mu} (Electric dipole moment)
    & \cite{molnet} & $\Lambda_1$ & 6,9 & 977 & 297 & 184 \\
    {\sc OptR} (Optical rotation)
    & \cite{pubchem} & $\Lambda_4$ & 5,44 & 147 & 107 & 55 \\
    {\sc Sl} (Solubility)
    & \cite{molnet} & $\Lambda_6$ & 4,55 & 915 & 300 & 175 \\
    {\sc SurfT} (Surface tension)
    & \cite{GDGPJDN.2017} & $\Lambda_7$ & 5,33 & 247 & 128 & 22 \\
    {\sc U0} (Internal energy at 0K)
    & \cite{molnet} & $\Lambda_1$ & 6,9 & 977 & 297 & 184\\
    {\sc Vd} (Vapor density)
    & \cite{pubchem} & $\Lambda_4$ & 4,30 & 474 & 214 & 53\\
    {\sc Visc} (Viscosity) 
    & \cite{GDPDGNVA.2020} & $\Lambda_7$ & 5,36 & 282 &126 & 22 \\
    {\sc Vp} (Vapor pressure)
    & \cite{pubchem} & $\Lambda_{2}$ & 4,5 & 482 & 238 & 96 \\
    \hline
  \end{tabular}
  }
\end{table}

\subsection{ML Experiments (Stage 3)}
For each property $\pi$,
we convert the data set $D_\pi$
into the set $f(D_\pi)$ of numerical vectors
by using a feature function $f:\MG\to\bbR^K$.
For $f$, we use $f=\fTL$ and
$f=f_{\textrm{2L+CC}}$, where $f_{\textrm{2L+CC}}$ is a feature function
such that $f_{\textrm{2L+CC}}(\bbC)=(\fTL(\bbC),\fCC(\bbC))$.
The purpose of the comparison is
to show that CC descriptors can extract 
useful information for ML. 
The number $K_{\textrm{CC}}$ of CC descriptors is
at most 70\% of the number $K_{\textrm{2L}}$ of 2L descriptors
for all data sets, as shown in \tabref{data_reg}. 

For $\pi$, let $D\subseteq D_\pi$ be a subset of the data set. 
To evaluate a prediction function $\eta:\bbR^K\to\bbR$ on $D$,
we employ
the determination of coefficient (R$^2$),
which is defined to be
\begin{align*}
  \textrm{R}^2(\eta,D)\triangleq1-\frac{\sum_{\bbC\in D}(a(\bbC)-\eta(f(\bbC)))^2}{\sum_{\bbC\in D}(a(\bbC)-\tilde{a})^2}\textrm{\ for\ }\tilde{a}=\frac{1}{|D|}\sum_{\bbC\in D}a(\bbC). 
\end{align*}


We construct prediction functions
based on Lasso linear regression (LLR)~\cite{Lasso96},
decision tree (DT)~\cite{DT.1986}
and random forest (RF)~\cite{RF.2001}. 
We evaluate the performance of each learning model  
by means of 10 repetitions of 5-fold cross validation.
Specifically, for each property $\pi$,
we divide the data set $D_\pi$ into 5 subsets randomly,
say $D_{\pi,1},\dots,D_{\pi,5}$,
so that $|D_{\pi,i}|-|D_{\pi,j}|\le1$ holds for $i,j=1,2,\dots,5$. 
For each $i=1,2,\dots,5$,
we construct a prediction function $\eta$ from
a subset $D_\pi\setminus D_{\pi,i}$ as the training set
and evaluate $\textrm{R}^2(\eta,D_{\pi,i})$
on the remaining subset $D_{\pi,i}$ as the test set. 
We take as the evaluation criterion
the median of $5\times 10=50$ values of R$^2$  
observed over 10 repetitions of 5-fold cross validation.  

We show the results in \tabref{R2}.
We may say that we can construct a good prediction function
for many data sets;
in 19 (resp., 11) out of the 27 data sets,
R$^2$ over 0.8 (resp., 0.9) is achieved.
We observe that, for some data sets,
there is a learning model that is not suitable.
For example, LLR attains poor performance for {\sc Vp},
regardless of feature functions, whereas
DT and RF are relatively good. 
\begin{table}[t]
  \centering
  \caption{ML results: medians of 50 values of R$^2$}
  \label{tab:R2}
  {\scriptsize
  \begin{tabular}{lc|rrr|rrr}
    \hline
    $\pi$ &\ \ \ & \multicolumn{3}{c|}{2L model $(f=\fTL)$} & \multicolumn{3}{c}{2L$+$CC model $(f=f_{\textrm{2L+CC}})$}\\
    && \multicolumn{1}{c}{LLR}
    & \multicolumn{1}{c}{DT}
    & \multicolumn{1}{c|}{RF}
    & \multicolumn{1}{c}{LLR}
    & \multicolumn{1}{c}{DT}
    & \multicolumn{1}{c}{RF}\\
    \hline 
    {\sc Alpha}  &&\ul{.961} &.769 &.856 & \ul{.961} & .784 & .875\\ 
    {\sc At}     &&.388 &.368 &.380 & \ul{.405} & .379 & \bf.401\\
    {\sc Bhl}    &&.483 &.401 &\ul{.555} & \bf.515 & *\bf.505 & \ul{.555}\\ 
    {\sc Bp}     &&.663 &.729 &.805 & .\bf701 & .728 & \ul{.824}\\ 
    {\sc Cv}     &&.970 &.805 &.911 & \ul{.979} & \bf.854 & .911\\ 
    {\sc Dc}     &&.574 &.408 &.624 & \bf.607 & *\bf.476 & \ul{.629}\\
    {\sc EDPA}   &&\ul{.999} &\ul{.999} &\ul{.999} & \ul{.999} & \ul{.999} & \ul{.999}\\ 
    {\sc Fp}     &&.570 &.572 &.748 & .564 & *\bf.645 & \ul{.752}\\
    {\sc Gap}    &&.783 &.668 &.733 & .776 & \bf.712 & *\ul{\bf.786}\\
    {\sc Hc}     &&\ul{\bf.951} &.826 &.894 & .924 & \bf.857 & .894\\ 
    {\sc Homo}   &&\ul{.707} &.391 &.556 & .703 & \bf.434 & *\bf.630\\ 
    {\sc Hv}  &&$-$13.744 &.128 &-0.058 & *\ul{\bf.817} & *\bf.554 & *\bf$-$0.001\\ 
    {\sc IhcLiq} &&\ul{.986} &.941 &.961 & \ul{.986} & .948 & .963\\ 
    {\sc IhcSol} &&.981 &.903 &.952 & \ul{.983} & .908 & .954\\ 
    {\sc KovRI}  &&.676 &.352 &.688 & *\ul{\bf.735} & *\bf.644 & .688\\ 
    {\sc Kow}    &&.952 &.854 &.911 & \ul{.960} & .871 & .923\\ 
    {\sc Lp}     &&.840 &.598 &.756 & \ul{.855} & .616 & \bf.796\\ 
    {\sc Lumo}   &&.841 &.734 &.796 & .836 & \bf.759 & \ul{\bf.842}\\ 
    {\sc Mp}     &&.785 &.687 &.805 & *\bf.836 & \bf.709 & \ul{\bf.839}\\ 
    {\sc mu}     &&.365 &.351 &.433 & .368 & \bf.400 & \ul{\bf.457}\\ 
    {\sc OptR}   &&.822 &.846 &\bf.891 & *\ul{\bf.933} & .861 & .871\\ 
    {\sc Sl}     &&.808 &.783 &.858 & .817 & .791 & \ul{.873}\\ 
    {\sc SurfT}  &&.803 &.645 &\ul{.840} & .809 & *\bf.714 & \ul{.840}\\ 
    {\sc U0}     &&\ul{.999} &.847 &.932 & \ul{.999} & *\bf.910 & .932 \\ 
    {\sc Vd}     &&.927 &.924 &.933 & .927 & \ul{.934} & \ul{.934}\\ 
    {\sc Visc}&&.893 &.860 &.909 & .894 & .866 & \ul{.910}\\ 
    {\sc Vp}   &&$-$0.013 &.771 &.857 & *\bf.115 & *\bf.845 & \ul{.861} \\ 
    \hline
  \end{tabular}
  }
\end{table}

Let us compare two feature functions, $\fTL$ and $f_{\textrm{2L+CC}}$.
For each property $\pi$, an underlined value indicates the maximum over the 6 values
($=$ 2 feature functions by 3 learning models).
The maximum is achieved for only 5 properties when $f=\fTL$,
whereas it is up to 25 properties when $f=f_{\textrm{2L+CC}}$. 
A bold-face (resp., *) indicates an R$^2$ value
that is larger at least by 0.02 (resp., 0.05)
than the R$^2$ value achieved by the other feature function and the same learning model. 
For example, for {\sc At},
the R$^2$ value 0.401 for $f_{\textrm{2L+CC}}$ and RF is bold
since it is larger than 0.379 for  $f_{\textrm{2L}}$ and RF by
$0.401-0.379=0.022>0.02$.
A bold value (resp., *)
appears only twice (resp., nowhere)
when $f=\fTL$, whereas it appears 31 (resp., 16) times when $f=f_{\textrm{2L+CC}}$.

We conclude that, in the 2L model,
the learning performance of
a prediction function can be improved by introducing CC descriptors. 

\subsection{Inference Experiments (Stage 4)}

For a property,
deciding two reals $\underline{y}^\ast,\overline{y}^\ast\in\bbR$ and
a specification $\sigma$,
we solve the MILP for inferring a chemical graph $\bbC^\dagger$.
Recall that the MILP consists of two families of constraints, that is
$(\MC_1)$ $\underline{y}^\ast\le \eta(x)\le\overline{y}^\ast$ and $x=f(\bbC^\dagger)$;
and $(\MC_2)$ $\bbC\in\MG_\sigma$.
In this experiment, we employ a hyperplane that is learned by LLR
for the prediction function $\eta$.
A hyperplane is a prediction function that is
represented by a pair $(w,b)\in\bbR^K\times\bbR$
and predicts the property value of a feature vector $x\in\bbR^K$
by $w(1)x(1)+\dots+w(K)x(K)+b$.
Hence, the constraint $\underline{y}^\ast\le \eta(x)\le\overline{y}^\ast$ in $\MC_1$
is represented by
\[
\underline{y}^\ast\le\sum_{j=1}^Kw(j)x(j)+b\le\overline{y}^\ast, 
\]
where use of a hyperplane in \molinf\ was
proposed in \cite{LLR.2022}. 
For the other constraints, see \appref{MILP}. 

\begin{figure}[t]
  \centering
  \begin{tabular}{ccccc}
    \includegraphics[width=2.5cm]{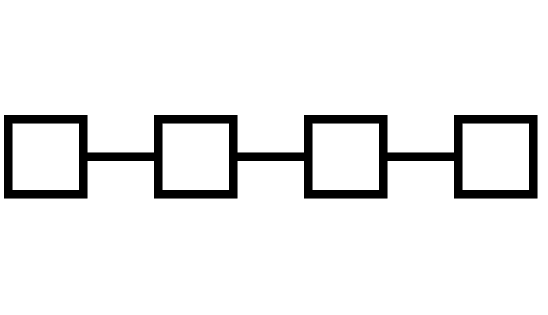} &
    \includegraphics[width=2.5cm]{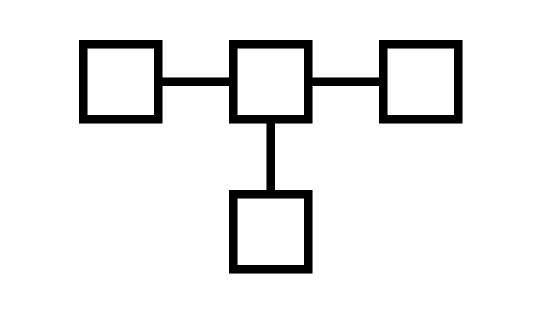}&
    \includegraphics[width=2.5cm]{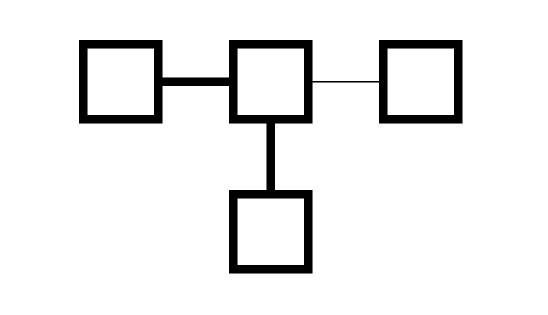}&
    \includegraphics[width=2.5cm]{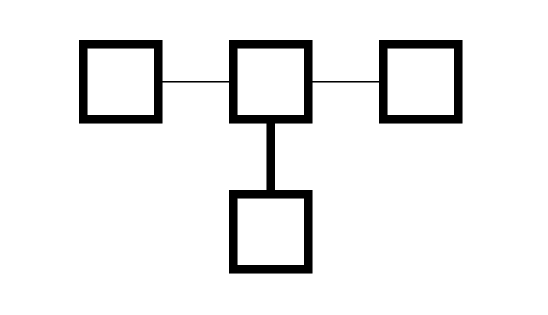}&
    \includegraphics[width=2.5cm]{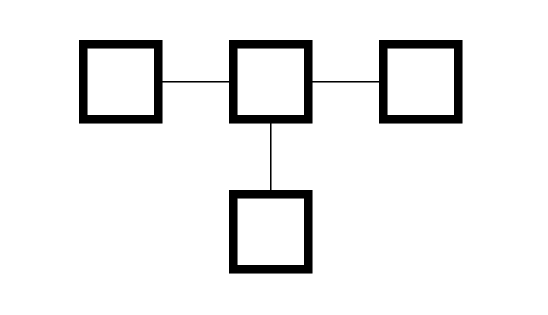}\\
    $\MT_{4\rma}$ & 
    $\MT_{4\rmb,0}$ & 
    $\MT_{4\rmb,1}$ & 
    $\MT_{4\rmb,2}$ & 
    $\MT_{4\rmb,3}$ 
  \end{tabular}

  \ \\

  \ \\
  
  \begin{tabular}{cccc}
    \includegraphics[width=3.2cm]{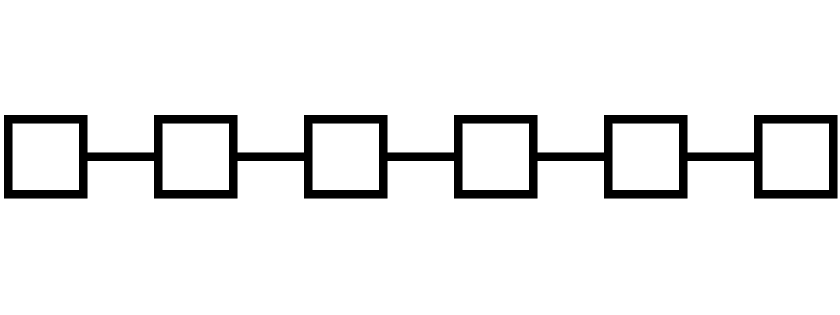} &
    \includegraphics[width=3.2cm]{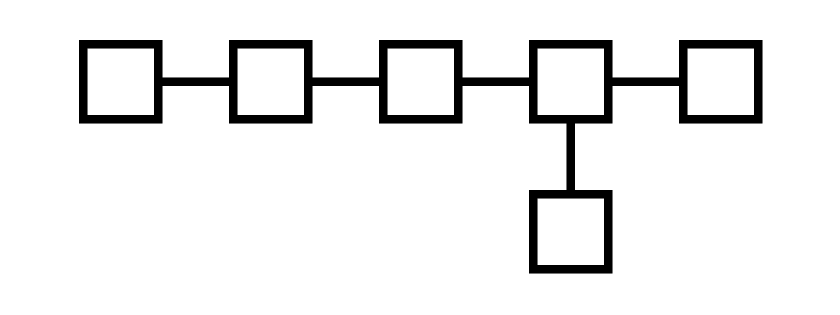}&
    \includegraphics[width=3.2cm]{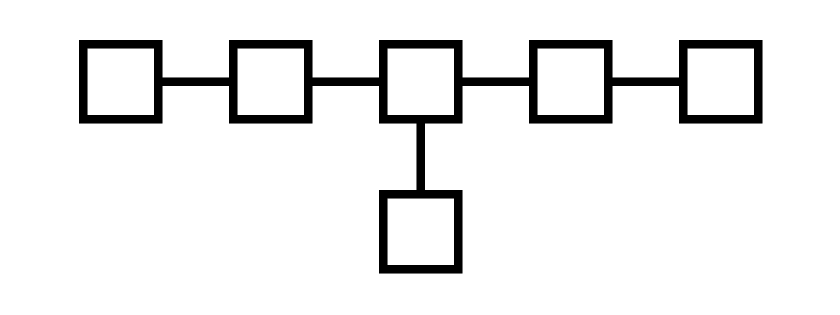}&
    \includegraphics[width=3.2cm]{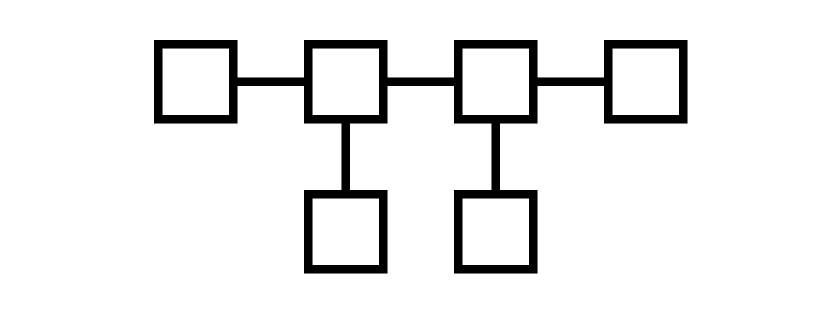}\\
    $\MT_{6\rma}$ & 
    $\MT_{6\rmb}$ & 
    $\MT_{6\rmb}$ & 
    $\MT_{6\rmd}$
  \end{tabular}
  \caption{Seed trees for the inference experiments: All nodes are ring nodes. A ring edge (resp., a non-ring edge) is depicted by a thick (resp., thin) line.}
  \label{fig:seed_tree_exp}
\end{figure}

We take up two properties {\sc Kow} and {\sc OptR},
for which LLR achieves R$^2$ over 0.9.
We consider 10 specifications
that have seed trees non-isomorphic to each other, where 
9 out of the 10 seed trees are shown in \figref{seed_tree_exp}. 
These seed trees are introduced to observe how computation time changes
with respect to the number of nodes; 
the number of ring edges; and the tree structure. 
The last seed tree is the one in \figref{seed_tree}(a). 
We denote this seed tree by $\MT_{5^\ast}$.
In each of the 10 specifications, 
we set other parameters  (see \tabref{spec}) than the seed tree sufficiently large
to the extent of the data set $D_\pi$. 
For example, we set $\MF^u:=\MF(D_\pi)$ for every $u\in T$,
that is, all fringe trees that appear in $D_\pi$ are available to $u$. 

We solve the MILP by utilizing {\tt CPLEX}~\cite{CPLEX} version 22.1.1.0.
We summarize statistics in Tables~\ref{tab:MILP_kow} and \ref{tab:MILP_optr}. 
The meanings of columns in the tables are described as follows.  
\begin{itemize}
\item \#V and \#C: the number of variables and constraints in MILP, respectively.
\item IP time: the computation time taken to solve the MILP.
\item $n(G^\dagger)$: the number of non-hydrogen atoms in the inferred chemical graph $G^\dagger$. 
\item $\eta(f(G^\dagger))$: an estimated property value of $G^\dagger$
  given by 
  the prediction function $\eta$ and
  the feature vector $f(G^\dagger)$, where $f=f_{\textrm{2L+CC}}$.  
\end{itemize}


\begin{table}[t]
  \centering
  \caption{Statistics of MILPs for {\sc Kow}: $\underline{y}^\ast$ and
    $\overline{y}^\ast$ are set to $-$7.53 and 15.60, respectively.}
  \label{tab:MILP_kow}
  \begin{tabular}{lcrrrrr}
    \hline
    Seed tree
    & \ \ \ %
    & \multicolumn{1}{c}{\#V}
    & \multicolumn{1}{c}{\#C}
    & \multicolumn{1}{c}{IP time (s)}
    & \multicolumn{1}{c}{$n(G^\dagger)$}
    & \multicolumn{1}{c}{$\eta(f(G^\dagger))$}\\
    \hline
    ${\mathcal T}_{\textrm{4a}}$ && 15139 & 18145 &5.2 & 18 &3.30\\
    ${\mathcal T}_{\textrm{4b,0}}$ && 15139 & 18145 & 4.6  & 18 & 3.64\\
    ${\mathcal T}_{\textrm{4b,1}}$ && 14921 & 13843 & 4.1 & 18 & 3.64\\
    ${\mathcal T}_{\textrm{4b,2}}$ && 14703 & 9541 & 38.4  & 25 & $-$5.38 \\
    ${\mathcal T}_{\textrm{4b,3}}$ && 14485 & 5239 & 7.0 & 44 &  0.19 \\
    ${\mathcal T}_{\textrm{6a}}$ && 21743 & 27843 & 7.8 & 26 & 4.42 \\
    ${\mathcal T}_{\textrm{6b}}$ && 21743 & 27843 & 11.1 & 26 & 4.76 \\
    ${\mathcal T}_{\textrm{6c}}$ && 21743 & 27843 & 9.4 & 26 & 4.76\\
    ${\mathcal T}_{\textrm{6d}}$ && 21743 & 27843 & 7.9 & 26 & 5.09 \\
    ${\mathcal T}_{5^\ast}$ && 19923 &10329 & 59.0 & 46 & $-$2.81\\
    \hline
  \end{tabular}
\end{table}

\begin{table}[t]
  \centering
  \caption{Statistics of MILPs for {\sc OptR}: $\underline{y}^\ast$ and
    $\overline{y}^\ast$ are set to $-$117.0 and 165.0, respectively.}
  \label{tab:MILP_optr}
  \begin{tabular}{lcrrrrr}
    \hline
    Seed tree
    & \ \ \ %
    & \multicolumn{1}{c}{\#V}
    & \multicolumn{1}{c}{\#C}
    & \multicolumn{1}{c}{IP time (s)}
    & \multicolumn{1}{c}{$n(G^\dagger)$}
    & \multicolumn{1}{c}{$\eta(f(G^\dagger))$}\\
    \hline
    ${\mathcal T}_{\textrm{4a}}$ && 8229 & 11853 & 12.1 & 21 & $-$53.06\\
    ${\mathcal T}_{\textrm{4b,0}}$ && 8229 & 11853 & 13.0 & 25 & $-$102.25 \\
    ${\mathcal T}_{\textrm{4b,1}}$ && 8122 & 9327 & 8.2 & 28 & $-$48.79 \\
    ${\mathcal T}_{\textrm{4b,2}}$ && 8015 & 6801 &28.8 & 30 & 45.47\\
    ${\mathcal T}_{\textrm{4b,3}}$ && 7908 & 4275 & 13.1 & 25 & 26.16\\
    ${\mathcal T}_{\textrm{6a}}$ && 11587 & 17875 & 13.7 & 34 & $-$54.39 \\
    ${\mathcal T}_{\textrm{6b}}$ && 11587 & 17875 & 27.7 & 34 & 134.51\\
    ${\mathcal T}_{\textrm{6c}}$ && 11587 & 17875 & 15.9 & 21 & $-$92.59\\
    ${\mathcal T}_{\textrm{6d}}$ &&  11587 & 17875 & 117.8 & 37 & $-$110.17\\
    ${\mathcal T}_{5^\ast}$ && 10652 & 7527 & 48.7 & 36 & $-$103.90\\
    \hline
  \end{tabular}
\end{table}


As shown in Tables~\ref{tab:MILP_kow} and \ref{tab:MILP_optr},
we can find chemical graphs
with up to 50 non-hydrogen atoms in a practical time;
the computation time is at most two minutes. 
There is a tendency such that
computation time is longer when there are more
\#V/\#C (i.e., the numbers of variables/constraints in MILP)
with some exceptions.
For example, for {\sc Kow}, the case of $\MT_{\textrm{4b,2}}$
takes 38.4 seconds, which is much more than
the cases where there are six ring nodes. 
Concerning \#V/\#C,
the more the ring nodes, the more they become.
The \#V/\#C are equal between seed trees
if they have the same numbers of ring nodes/edges;
e.g., \#V/\#C are equal between $\MT_{\textrm{4a}}$ and $\MT_{\textrm{4b,0}}$. 

\begin{figure}[t]
  \centering
  \begin{tabular}{cccc}
    \includegraphics[width=2.5cm]{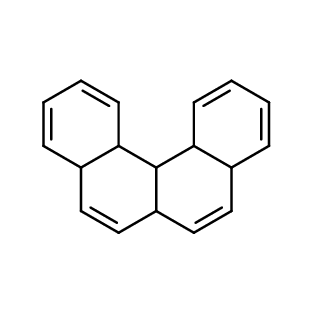}&
    \includegraphics[width=2.5cm]{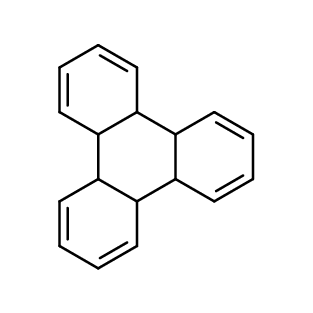}&
    \includegraphics[width=2.5cm]{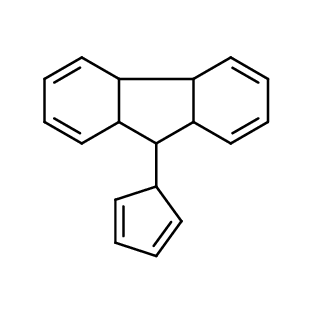}&
    \includegraphics[width=4cm]{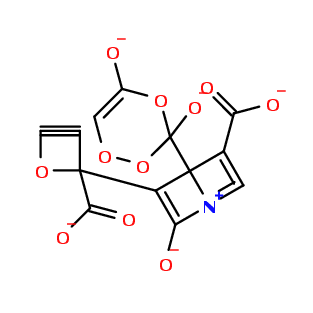}\\
    $\MT_\textrm{4a}$ ({\sc Kow})
    &$\MT_\textrm{4b,0}$ ({\sc Kow})
    &$\MT_\textrm{4b,1}$ ({\sc Kow})
    &$\MT_\textrm{4b,2}$ ({\sc Kow})
  \end{tabular}

  \begin{tabular}{cc}
    \includegraphics[width=6cm]{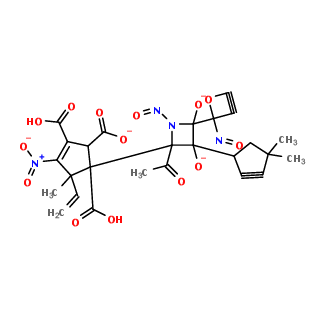}&
    \includegraphics[width=6cm]{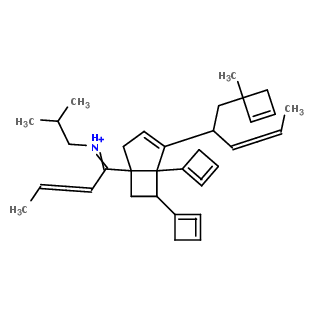}\\
    $\MT_\textrm{4b,3}$ ({\sc Kow})
    &$\MT_{5^\ast}$ ({\sc OptR})
  \end{tabular}
  \caption{Inferred chemical graphs}
  \label{fig:MILP}
\end{figure}

We also show some of the inferred chemical graphs in \figref{MILP}.
As expected, ring nodes in the seed trees
are expanded to cycles in the chemical graphs.
Some graphs contain 4-cycles or ionized elements.
We can prevent MILP from using such structures
by setting specifications appropriately.

\section{Concluding Remarks}
\label{sec:conc}

In this paper, we proposed a new family of descriptors,
cycle-configurations, that can be used in the standard 2L model of \molinf. 
We introduced the definition in \secref{cc} and described
how we deal with them in the MILP 
in \secref{milp}. 
Then in \secref{exp},
we demonstrated that the performance of a prediction function
is improved in many cases when we introduce CC descriptors. 
We also showed that a chemical graph with up to 50 non-hydrogen atoms
can be inferred in a practical time. 

The 2L+CC model can be extended further in the similar way as the 2L model. 
Specifically, we can enumerate isomers of the inferred graph 
by dynamic programming~\cite{Zhu.2021aa}\ %
or generate ``close'' compounds in the sense of property values
by a grid neighborhood approach~\cite{gridAZHZNA21}.
Note that the constraint $\MC_1$ of the MILP
can contain multiple prediction functions for multiple properties,
as is done in \cite{LLR.2022},
where we have included a single property
in this paper for simplicity. 
Besides, we may apply the 2L+CC model to inference of polymers. 
These are left for future work. 


\clearpage
\bibliographystyle{abbrv}
\bibliography{CC_ref}

\clearpage
\appendix

\newpage
\section*{Appendix}

\section{A Full Description of Descriptors in the 2L Model}
\label{app:2Lmodel}
 
Associated with the two functions 
$\alpha$ and $\beta$ in a chemical graph $\bbC=(H,\alpha,\beta)$,
we introduce functions  
 $\ac: V(E)\to (\Lambda\setminus\{\ttH\})\times (\Lambda\setminus\{\ttH\})\times [1,3]$, 
 $\cs: V(E)\to (\Lambda\setminus\{\ttH\})\times [1,4]$ and
$\ec: V(E)\to ((\Lambda\setminus\{\ttH\})\times [1,4])\times ((\Lambda\setminus\{\ttH\})\times [1,4])\times [1,3]$
in the following. 

 To represent  a feature of the exterior  of  $\bbC$, 
  a  chemical rooted tree in $\mathcal{T}(\bbC)$ is
  called a {\em fringe-configuration} of $\bbC$. 

We also represent leaf-edges in the exterior of $\bbC$.
For a leaf-edge $uv\in E(\hsup{\bbC})$ with $\deg_{\hsup{\bbC}}(u)=1$, we define
the {\em adjacency-configuration} of $e$ to be an ordered tuple
$(\alpha(u),\alpha(v),\beta(uv))$. 
Define 
\[ \Gac^\lf\triangleq \{(\tta,\ttb,m)\mid \tta,\ttb\in\Lambda, 
m\in[1,\min\{\val(\tta),\val(\ttb)\}]\} \]
as a set of possible adjacency-configurations for leaf-edges. 

To  represent a feature of an interior-vertex $v\in V^\inte(\bbC)$ such that
$\alpha(v)=\tta$  and  $\deg_{\hsup{\bbC}}(v)=d$
(i.e., the number of non-hydrogen atoms adjacent to $v$ is $d$) 
   in a chemical   graph  $\bbC=(H,\alpha,\beta)$,
 we use  a pair $(\tta, d)\in (\Lambda\setminus\{{\tt H}\})\times [1,4]$,
 which we call the {\em chemical symbol} $\cs(v)$ of the vertex $v$.
 We treat $(\tta, d)$ as a single symbol $\tta d$,  and  
define $\Ldg$   to be  the set of all chemical symbols
$\mu=\tta d\in  (\Lambda\setminus\{{\tt H}\})\times [1,4]$.  

We define a method for featuring interior-edges  as follows.
Let $e=uv\in E^\inte(\bbC)$  be 
 an interior-edge $e=uv\in E^\inte(\bbC)$ 
 such that $\alpha(u)=\tta$, $\alpha(v)=\ttb$ and $\beta(e)=m$ 
   in a chemical graph  $\bbC=(H,\alpha,\beta)$.
To feature this edge $e$, 
 we use a tuple $(\tta,\ttb,m)\in (\Lambda\setminus\{{\tt H}\})
    \times (\Lambda\setminus\{{\tt H}\})\times [1,3]$,
 which we call the {\em adjacency-configuration} $\ac(e)$ of the edge $e$. 
 We introduce a total order $<$ over the elements in $\Lambda$
 to distinguish  between $(\tta,\ttb, m)$ and $(\ttb,\tta, m)$ 
 $(\tta\neq \ttb)$ notationally.
 For a tuple  $\nu=(\tta,\ttb, m)$,
 let $\overline{\nu}$ denote the tuple $(\ttb,\tta, m)$.

Let $e=uv\in E^\inte(\bbC)$  be 
an  interior-edge $e=uv\in E^\inte(\bbC)$ 
 such that $\cs(u)=\mu$, $\cs(v)=\mu'$ and $\beta(e)=m$ 
   in a chemical  graph  $\bbC=(H,\alpha,\beta)$.
To feature this edge $e$, 
 we use a tuple $(\mu,\mu',m)\in \Ldg\times \Ldg\times [1,3]$, 
 which we call  the {\em edge-configuration} $\ec(e)$ of the edge $e$. 
 We introduce a total order $<$ over the elements in $\Ldg$
 to distinguish between $(\mu,\mu', m)$ and $(\mu', \mu, m)$ 
 $(\mu \neq \mu')$ notationally. 
 For a tuple  $\gamma=(\mu,\mu', m)$,
 let $\overline{\gamma}$ denote the tuple $(\mu', \mu, m)$. 

Let $\pi$ be a chemical property for which we will construct
a prediction function $\eta$ from a feature
vector $f(\bbC)$ of a chemical graph $\bbC$ 
to a predicted value $y\in \mathbb{R}$
for the  chemical property of $\bbC$.

We first choose a set $\Lambda$ of chemical elements
 and then collect a data set  $D_{\pi}$ of
  chemical compounds  $C$ whose 
  chemical elements belong to $\Lambda$,
  where we regard  $D_{\pi}$ as a set of chemical graphs $\bbC$
  that represent the chemical compounds $C$  in  $D_{\pi}$.
To define the interior/exterior of 
chemical graphs  $\bbC\in D_{\pi}$,
we  next choose a branch-parameter ${\rho}$, where
 we recommend ${\rho}=2$.  
 
Let $\Lambda^\inte(D_\pi)\subseteq \Lambda$ 
(resp., $\Lambda^\ex(D_\pi)\subseteq \Lambda$)
denote the set  of chemical elements  used in
the set $V^\inte(\bbC)$ of interior-vertices
(resp., the set $V^\ex(\bbC)$ of  exterior-vertices) of $\bbC$
 over all chemical graphs $\bbC\in D_\pi$, 
and $\Gamma^\inte(D_\pi)$
denote the set of edge-configurations used in
the set $E^\inte(\bbC)$  of interior-edges in $\bbC$
 over all chemical graphs $\bbC\in D_\pi$. 
Let $\mathcal{F}(D_\pi)$ denote the set of
chemical rooted trees $\psi$  
r-isomorphic to a chemical rooted tree in $\mathcal{T}(\bbC)$
  over all chemical graphs $\bbC\in D_\pi$,
  where possibly a chemical rooted tree $\psi\in \mathcal{F}(D_\pi)$
  consists of a single chemical element $\tta\in \Lambda\setminus \{{\tt H}\}$.
  
We define an integer encoding of a finite set $A$ of elements
to be a bijection $\pi: A \to [1, |A|]$, 
where we denote by $[A]$   the set $[1, |A|]$ of integers.
Introduce  an integer coding of each of the   sets 
$\Lambda^\inte(D_\pi)$, $\Lambda^\ex(D_\pi)$, 
$\Gamma^\inte(D_\pi)$ and $\mathcal{F}(D_\pi)$. 
Let $[\tta]^\inte$  
(resp., $[\tta]^\ex$)  denote   
the coded integer of  an element $\tta\in \Lambda^\inte(D_\pi)$
(resp., $\tta\in \Lambda^\ex(D_\pi)$),  
$[\gamma]$   denote  
the coded integer of  an element $\gamma$ in $\Gamma^\inte(D_\pi)$
and 
$[\psi]$   denote  an element $\psi$ in $\mathcal{F}(D_\pi)$. 
 
 Over 99\% of  chemical compounds $\bbC$ with up to
  100 non-hydrogen atoms in  PubChem  have degree at most 4
 in the hydrogen-suppressed graph $\hsup{\bbC}$~\cite{AZSSSZNA20}. 
We assume that a chemical graph $\bbC$
 treated in this paper satisfies  $\deg_{\hsup{\bbC}}(v)\leq 4$
in the hydrogen-suppressed graph $\hsup{\bbC}$.
 
In our model, we  use an integer 
  $\mathrm{mass}^*(\tta)=\lfloor 10\cdot \mathrm{mass}(\tta)\rfloor$, 
 for each $\tta\in \Lambda$.
 
 For a chemical property $\pi$,
 we define a set $D_\pi^{(1)}$ of descriptors  
  of a  chemical graph $\bbC=(H,\alpha,\beta)\in D_{\pi}$ 
  to be  the following  
non-negative values $\dcp_i(\bbC)$, $i\in [1,K_{\textrm{2L}}]$, where 
$K_{\textrm{2L}}= 14+ |\Lambda^\inte(D_\pi)|+|\Lambda^\ex(D_\pi)|
         +|\Gamma^\inte(D_\pi)|+|\mathcal{F}(D_\pi)|+|\Gac^\lf|$. 


\begin{enumerate}  
\item   
$\dcp_1(\bbC)$: the number  $|V(H)|-|V_{\ttH}|$ of non-hydrogen atoms  in  $\bbC$.  
 
\item   
$\dcp_2(\bbC)$: the rank 
of   $\bbC$ (i.e., the minimum number of edges to be removed to make the graph 
acyclic).  

\item 
$\dcp_3(\bbC)$:  the number $|V^\inte(\bbC)|$ of interior-vertices in  $\bbC$.
  
\item 
$\dcp_4(\bbC)$: 
the average $\overline{\mathrm{ms}}(\bbC)$ of mass$^*$ 
over all atoms in $\bbC$; \\
 i.e., $\overline{\mathrm{ms}}(\bbC)\triangleq 
 \frac{1}{|V(H)|}\sum_{v\in V(H)}\mathrm{mass}^*(\alpha(v))$. 

\item 
$\dcp_i(\bbC)$,  $i=4+d,   d\in [1,4]$: 
the number $\dg_d^{\oH} (\bbC)$ 
 of non-hydrogen vertices $v\in V(H)\setminus V_{\ttH}$
 of degree $\deg_{\hsup{\bbC}}(v)=d$
 in the hydrogen-suppressed chemical graph $\hsup{\bbC}$.  
 
\item 
$\dcp_i(\bbC)$,  $i=8+d,   d\in [1,4]$: 
the number $\dg_d^\inte(\bbC)$
 of interior-vertices of interior-degree  $\deg_{\bbC^\inte}(v)=d$
  in the interior $\bbC^\inte=(V^\inte(\bbC),E^\inte(\bbC))$ of  $\bbC$. 
  
   
\item $\dcp_i(\bbC)$, $i=12+m$,  $m\in[2,3]$: 
the number $\bd_m^\inte(\bbC)$
 of  interior-edges with bond multiplicity $m$ in  $\bbC$; 
 i.e., $\bd_m^\inte(\bbC)\triangleq |\{e\in E^\inte(\bbC)\mid \beta(e)=m\}|$.

\item $\dcp_i(\bbC)$, $i=14+[\tta]^\inte$, 
 $\tta\in \Lambda^\inte(D_\pi)$: 
 the frequency $\na_\tta^\inte(\bbC)=|V_\tta(\bbC)\cap V^\inte(\bbC) |$ 
 of chemical element $\tta$ in
 the set $V^\inte(\bbC)$ of  interior-vertices in  $\bbC$. 
 
\item $\dcp_i(\bbC)$, 
$i=14+|\Lambda^\inte(D_\pi)|+[\tta]^\ex$, 
 $\tta\in \Lambda^\ex(D_\pi)$: 
 the frequency $\na_\tta^\ex(\bbC)=|V_\tta(\bbC)\cap V^\ex(\bbC) |$
  of chemical element $\tta$ in
 the set $V^\ex(\bbC)$ of  exterior-vertices in  $\bbC$. 
 
\item $\dcp_i(\bbC)$, 
$i=14+|\Lambda^\inte(D_\pi)|+|\Lambda^\ex(D_\pi)|+ [\gamma]$, 
$\gamma \in \Gamma^\inte(D_\pi)$: 
the frequency $\ec_{\gamma} (\bbC)$ of edge-configuration $\gamma$
in the set $E^\inte(\bbC)$ of interior-edges   in  $\bbC$.

\item $\dcp_i(\bbC)$, 
$i= 14+|\Lambda^\inte(D_\pi)|+|\Lambda^\ex(D_\pi)|
+ |\Gamma^\inte(D_\pi)|+ [\psi]$,  
 $\psi \in \mathcal{F}(D_\pi)$: 
the frequency $\fc_{\psi}(\bbC)$ of fringe-configuration $\psi $
in the set of ${\rho}$-fringe-trees in  $\bbC$. 

\item $\dcp_i(\bbC)$, 
$i= 14+|\Lambda^\inte(D_\pi)|+|\Lambda^\ex(D_\pi)|
+ |\Gamma^\inte(D_\pi)|+|\mathcal{F}(D_\pi)|+ [\nu]$,  
 $\nu \in \Gac^\lf$: 
the frequency $\ac_{\nu}^\lf(\bbC)$ of adjacency-configuration $\nu$
in the set of leaf-edges in  $\hsup{\bbC}$. 
\end{enumerate}

\section{MILP Formulation for the 2L$+$CC Model}
\label{app:MILP}

Let $\MT=(T;\Vcirc,\Ecirc)$ denote a seed tree. 
Each ring node $u\in\Vcirc$ is expanded to a cycle
whose length is between $\Cmin$ and $\Cmax$. 
This expansion is done by assigning
$\xi\in\Xi$ to $u$, where $\Xi^u$ is the set of cycle-configurations
available to $u$
such that every $\xi\in\Xi$ satisfies $\Cmin\le|\xi|\le\Cmax$. 
Strictly speaking,
a ring node $u\in\Vcirc$ is assigned a graph $C^u$
such that $V(C^u)=\{u_1,u_2,\dots,u_{\Cmax}\}$ and
$E(C^u)=\{e^u_1,e^u_2,\dots,e^u_{2\Cmax-\Cmin}\}$,
where, for $i\in[1,2\Cmax-\Cmin]$,
\[
e^u_i=\left\{
\begin{array}{ll}
  u_iu_{i+1} & \textrm{if\ }i\le\Cmax,\\
  u_1u_{i-\Cmax+\Cmin-1} & \textrm{otherwise},
\end{array}
\right.
\]
and we regard $u_{\Cmax+1}=u_1$ for convenience.
The vertices and edges that form the cycle
are chosen according to $|\xi|$, where $\xi$ is the cycle-configuration
assigned to $u$.
Specifically, vertices $u_1,u_2,\dots,u_{|\xi|}$
and edges $u_1u_2,\dots,u_{|\xi|-1}u_{|\xi|},u_{|\xi|}u_1$
are chosen.

%
%
For a cycle-configuration $\xi$ and $r\in[1,\Cmax]$, 
let us define
\begin{align*}
  \hat{\xi}^{+}(r)\triangleq\{(\mu,\mu_0)\mid \mu,\mu_0\in\{u_1,\dots,u_{|\xi|}\},\ \xi(\mu-\mu_0+1)=r\};\\
  \hat{\xi}^{-}(r)\triangleq\{(\mu,\mu_0)\mid \mu,\mu_0\in\{u_1,\dots,u_{|\xi|}\},\ \xi(\mu_0-\mu+1)=r\},
\end{align*}
and for $\mu\in V(C^u)$ and $\delta\in\{+,-\}$,
we let
\[
\hat{\xi}^{\delta}(r,\mu):=\{\mu_0\in V(C^u)\mid (\mu,\mu_0)\in\hat{\xi}^{\delta}(r)\}.
\]


\subsection{Assigning Cycle-Configurations to Ring Nodes}
{\bf Constants.}
\begin{itemize}
\item A seed tree $\MT=(T;\Vcirc,\Ecirc)$;
\item  the set $\Xi^u$ of available cycle-configurations
  for each $u\in \Vcirc$, $\Xi^\MT := \bigcup_{u \in \Vcirc} \Xi^u$;
\item a positive constant $\varepsilon_1\in\mathbb{R}_+$
  that represents a sufficiently small number;
\item a positive constant $M_1\in\mathbb{R}_+$
  that represents a sufficiently large number. 
\end{itemize}

\noindent
{\bf Variables.}
\begin{itemize}
\item Real variables $y^u_{[\mu]}$, $u\in \Vcirc$, $\mu\in V(C^u)$
  that store the mass sum in the fringe-tree
  attached to vertex $\mu\in V(C^u)$;
\item real variables $z^u_r$, $u\in \Vcirc$, $r\in[1,\Cmax]$ that
  represent the $r$-th smallest mass sum of a fringe-tree in $C^u$;
\item binary variables $x^u_{[\xi],[\mu_0],\delta}$, $u\in \Vcirc$,
  $\xi\in\Xi^u$,
  $\mu_0\in\{u_1,\dots,u_{|\xi|}\}$ and $\delta\in\{+,-\}$,
  indicating whether $\xi$ is
  assigned to the starting point $\mu_0$ in the direction $\delta$;
\item binary variables $x_{[\xi]}^u$, $u \in \Vcirc$, $\xi \in \Xi^u$, 
indicating whether $\xi$ is assigned to $C^u$;
\item integer variables $\cc([\xi])$, $\xi \in \Xi^\MT$, cycle-configurations.
\end{itemize}

\noindent
{\bf Constraints.}

\noindent
For each $u\in \Vcirc$:
\begin{align}
  &0\le z^u_1,\dots,z^u_{\Cmax}\le M_1,\\
  &z^u_r+\varepsilon_1\le z^u_{r+1},&&r\in[1,\Cmax-1];\\
    &x_{[\xi]}^u = \sum_{\mu_0\in\{u_1,\dots,u_{|\xi|}\}} \sum_{\delta\in\{+,-\}} x_{[\xi],[\mu_0],\delta}^u, && \xi \in \Xi^u;\\
  &\sum_{\xi\in\Xi^u}x_{[\xi]}^u=1;\\
  &y^u_{[\mu]}\le z^u_r+M_1\big(1-\sum_{\xi\in\Xi^u}\sum_{\delta\in\{+,-\}}\sum_{\mu_0\in\hat{\xi}^\delta(r,\mu)} x^u_{[\xi],[\mu_0],\delta}\big),&&
  r\in[1,\Cmax],\ \mu\in V(C^u);\\
  &y^u_{[\mu]}\ge z^u_r-M_1\big(1-\sum_{\xi\in\Xi^u}\sum_{\delta\in\{+,-\}}\sum_{\mu_0\in\hat{\xi}^\delta(r,\mu)} x^u_{[\xi],[\mu_0],\delta}\big),&&
  r\in[1,\Cmax],\ \mu\in V(C^u). 
\end{align}

\noindent
For each $\xi \in \Xi^\MT$:
\begin{align}
& \cc([\xi]) = \sum_{u \in \Vcirc, \xi \in \Xi^u} x_{[\xi]}^u , &&
\end{align}

\subsection{Associating Ring Nodes with Ring Edges}

\noindent
{\bf Variables.}
\begin{itemize}
\item Binary variables $e_i^u, u \in \Vcirc, i \in [1, 2\Cmax-\Cmin]$, indicating whether the edge $e_i$ in $C^u$ is used;
\item binary variables $x^{\re,u}_{[e],[\nu]}$, $u\in\Vcirc$,
  $e\in \Ecirc(u)$, $\nu\in E(C^u)$;
\item binary variables $x^{\rn,u}_{[e'],[\mu]}$, $u\in\Vcirc$,
  $e'\in \barEcirc(u)$, $\mu\in V(C^u)$; 
\end{itemize}

\noindent
{\bf Constraints.}

\noindent
For each $u\in \Vcirc$:

\begin{align}
  & \sum_{\nu\in E(C^u)} x^{\re,u}_{[e],[\nu]}=1, &&e\in\Ecirc(u);\\
  & \sum_{\mu\in V(C^u)} x^{\rn,u}_{[e'],[\mu]}=1, &&e'\in \barEcirc(u);
\end{align}


\noindent
\begin{align}
& \sum_{e \in \Ecirc(u)} \sum_{\nu \in (E(C^u))(\mu)} x_{[e], [\nu]}^{\re, u} \le 1, && \mu \in V(C^u); \\
& \sum_{e \in \Ecirc(u)} x_{[e], [\nu]}^{\re, u} \le 1, && \nu \in E(C^u);
\end{align}

For each $u \in \Vcirc$:
\begin{align}
& e_i^u = 1, && i \in [1, \Cmin - 1] ; \nonumber \\
& e_i^u = \sum_{\xi \in \Xi^u, |\xi| > i} x_{[\xi]}^u, && i \in [\Cmin, \Cmax-2]; \nonumber  \\
& e_i^u = \sum_{\xi \in \Xi^u, |\xi| = \Cmax} x_{[\xi]}^u, && i \in \{ \Cmax-1, \Cmax \}; \nonumber  \\
& e_i^u = \sum_{\xi \in \Xi^u, |\xi| = i - \Cmax+\Cmin-1} x_{[\xi]}^u, && i \in [\Cmax+1, 2\Cmax-\Cmin]; \\
& x_{[e], i}^{\re, u} \le e_i^u, && e \in \Ecirc(u), i \in [1, 2\Cmax-\Cmin];
\end{align}

%
%


\begin{align}
& x_{[e'],i}^{\rn, u} \le \sum_{\xi \in \Xi^u, |\xi| \ge i} x_{[\xi]}^u, & & e' \in \barEcirc(u), i \in [\Cmin+1, \Cmax];
\end{align}


\subsection{Constraints for Including Fringe-Trees}
For a leaf-edge $uv \in E(G_\MT)$ with $\deg_{G_\MT}(u)=1$, we define the \emph{adjacency-configuration} of $uv$ to be an ordered tuple $(\alpha(u),\alpha(v), \beta(uv))$.

\noindent
{\bf Constants.}
\begin{itemize}
\item The set $\fr^u$ of the available fringe-trees for each 
$u \in V(T)$,  $\fr^\MT := \bigcup_{u \in V(T)} \fr^u $;
\item the set $\Gaclf$ of available adjacency-configuration on the set of leaf-edges;
\item functions $\msfr(\psi), \htfr(\psi), \nnoH(\psi), \aclfgamma(\psi)$ denoting the mass, height, number of non-hydrogen non-root atoms, number of leaf-edge adjacency-configurations $\gamma$ of the fringe-tree $\psi$, respectively;
\item integers $\nLB, \nUB$, that represent the lower and upper bounds on the number of non-hydrogen atoms in $G_\MT$, respectively;
\item integers $\ninLB, \ninUB$, that represent the lower and upper bounds on the number of non-hydrogen atoms in the interior part of $G_\MT$, respectively;
\item integers $\fcLB(\psi), \fcUB(\psi) \in [0, \nUB], \psi \in \fr^\MT$, that represent the lower and upper bounds on the fringe-configurations, respectively;
\item integers $\aclfLB(\gamma), \aclfUB(\gamma) \in [0, \nUB], \gamma \in \Gaclf$, that represent the lower and upper bounds on the adjacency-configurations of leaf-edges, respectively;
\end{itemize}

\noindent
{\bf Variables.}
\begin{itemize}
\item Binary variables $\dfr(u, [\mu];[\psi])$, $u \in \Vcirc, \mu \in V(C^u)$, $\psi \in \fr^u$,
indicating whether the fringe-tree $\psi$ is attached to vertex $\mu \in V(C^u)$;
\item binary variables $\dfr(v;[\psi])$, $v \in V(T) \setminus \Vcirc$, $\psi \in \fr^v$,
indicating whether the fringe-tree $\psi$ is attached to node $v \in V(T) \setminus \Vcirc$;
\item integer variables $\fc([e];[\psi]) \in [0,2]$, $e \in \Ecirc$, $\psi \in \fr^\MT$, that stores the number of the fringe-tree $\psi$ is used in the ring edge $e \in \Ecirc$;
\item integer variable $\rank$ that represents the rank of $G_\MT$;
\item integer variables $n_G \in [\nLB, \nUB], \nint \in [\ninLB, \ninUB]$ that represents the number of non-hydrogen atoms in $G_\MT$ and the interior part of $G_\MT$, respectively;
\item integer variables $\fc([\psi]) \in [\fcLB(\psi), \fcUB(\psi)]$, $\psi \in \fr^\MT$ that stores the fringe-configurations;
\item integer variables $\aclf([\gamma]) \in [\aclfLB(\gamma), \aclfUB(\gamma)]$, $\gamma \in \Gaclf$, that stores the adjacency-configurations of leaf-edges;
\end{itemize}

\noindent
{\bf Constraints.}

\noindent
For each $u \in \Vcirc, \mu \in V(C^u)$:
\begin{align}
& \sum_{\psi \in \fr^u} \dfr(u, [\mu]; [\psi]) = \sum_{\xi \in \Xi^u, |\xi| \geq [\mu]} x_{[\xi]}^u; && \\
& \sum_{\psi \in \fr^u} \msfr(\psi) \cdot \dfr(u, [\mu]; [\psi]) = y_{[\mu]}^u; && 
\end{align}

\noindent
For each $v \in V(T) \setminus \Vcirc$:
\begin{align}
& \sum_{\psi \in \fr^v} \dfr(v; [\psi]) = 1,  && \nonumber \\ 
& \sum_{\psi \in \fr^v, \htfr(\psi) = \rho} \dfr(v; [\psi]) = 1,  &&  v \ \textrm{is\ a\ leaf\ of} \ T; 
\end{align}

\noindent
For each $e=uu'\in\Ecirc$ such that $[u]<[u']$:
\begin{align}
 \sum_{\psi \in \fr^u} [\psi] \cdot \dfr(u, i_1; [\psi]) - \sum_{\psi \in \fr^{u'}} [\psi] \cdot \dfr(u', j_2; [\psi]) \leq & |\fr^\MT| (2 - x^{\re,u}_{[e],[\nu]} - x^{\re,u'}_{[e],[\nu']}), \nonumber \\
 \sum_{\psi \in \fr^u} [\psi] \cdot \dfr(u, i_1; [\psi]) - \sum_{\psi \in \fr^{u'}} [\psi] \cdot \dfr(u', j_2; [\psi]) \geq & |\fr^\MT| (x^{\re,u}_{[e],[\nu]} + x^{\re,u'}_{[e],[\nu']} - 2), \nonumber \\
 \sum_{\psi \in \fr^u} [\psi] \cdot \dfr(u, i_2; [\psi]) - \sum_{\psi \in \fr^{u'}} [\psi] \cdot \dfr(u', j_1; [\psi]) \leq & |\fr^\MT| (2 - x^{\re,u}_{[e],[\nu]} - x^{\re,u'}_{[e],[\nu']}), \nonumber \\
 \sum_{\psi \in \fr^u} [\psi] \cdot \dfr(u, i_2; [\psi]) - \sum_{\psi \in \fr^{u'}} [\psi] \cdot \dfr(u', j_1; [\psi]) \geq & |\fr^\MT| (x^{\re,u}_{[e],[\nu]} + x^{\re,u'}_{[e],[\nu']} - 2), \nonumber \\
  & \nu=u_{i_1}u_{i_2}\in E(C^u),\ \nu'=u'_{j_1}u'_{j_2}\in E(C^{u'})\nonumber\\ 
  & \textrm{such\ that\ }i_1<i_2,\ j_1<j_2; 
\end{align}
\begin{align}
\fc([e]; [\psi]) - \dfr(u, i_1; [\psi]) - \dfr(u, i_2; [\psi]) \le 2(1 - x_{[e], [\nu]}^{\re, u}), & \nonumber  \\
\fc([e]; [\psi]) - \dfr(u, i_1; [\psi]) - \dfr(u, i_2; [\psi]) \ge 2( x_{[e], [\nu]}^{\re, u} - 1),  &\nu = u_{i_1}u_{i_2} \in E(C^u), \psi \in \fr^\MT;
\end{align}

\noindent
For each $\psi \in \fr^\MT$:
\begin{align}
& \fc([\psi]) = \sum_{u \in \Vcirc} \sum_{\mu \in V(C^u)} \dfr(u, [\mu]; [\psi]) + \sum_{v \in V(T) \setminus \Vcirc} \dfr(v; [\psi]) - \sum_{e \in \Ecirc} \fc([e];[\psi]); &&
\end{align}

\noindent
For each $\gamma \in \Gaclf$:
\begin{align}
& \aclf([\gamma]) = \sum_{\psi \in \fr^\MT} \aclfgamma(\psi) \fc([\psi]);
\end{align}

\begin{align}
& \rank = |\Vcirc|; && \\
& \nint = \sum_{u \in \Vcirc} \sum_{\xi \in \Xi^u} |\xi| \cdot x_{[\xi]}^u + |V(T) \setminus \Vcirc| - 2|\Ecirc|; && \\
& n_G = \nint + \sum_{\psi \in \fr^\MT} \nnoH(\psi)\fc([\psi]); &&
\end{align}

\subsection{Descriptors for the Number of Specified Degree}
{\bf Constants.}
\begin{itemize}
\item Function $\dgnoH(\psi)$ denoting the degree of the root of the fringe tree $\psi$;
\end{itemize}

\noindent
{\bf Variables.}
\begin{itemize}
\item Binary variables $\ddg(u, [\mu]; d), u \in \Vcirc, \mu \in V(C^u), d \in [1,4]$, indicating the degree of $\mu$ in $G_{\MT}$;
\item binary variables $\ddg(v; d), v \in V(T) \setminus \Vcirc, d \in [1,4]$,  indicating the degree of $v$ in $G_{\MT}$;
\item binary variables $\ddgint(u, [\mu]; d), u \in \Vcirc, \mu \in V(C^u), d \in [1,4]$, indicating the interior degree of $\mu$ in $G_{\MT}$;
\item binary variables $\ddgint(v; d), v \in V(T) \setminus \Vcirc, d \in [1,4]$,  indicating the interior degree of $v$ in $G_{\MT}$;
\item integer variables $\dg(d), d \in [1,4]$, that stores the number of vertices with degree $d$ in $G_\MT$;
\item integer variables $\dgint(d), d \in [1,4]$, that stores the number of vertices with interior degree $d$ in $G_\MT$;
\item integer variables $\dg([e];d) \in [0,2], e \in \Ecirc, d \in [1,4]$, that stores the number of vertices with degree $d$ in the ring edge $e$;
\item integer variables $\dgint([e];d) \in [0,2], e \in \Ecirc, d \in [1,4]$, that stores the number of vertices with interior degree $d$ in the ring edge $e$;
\item integer variables $\dg_{[e], [\mu], +}^{\re, u}, u \in \Vcirc, \mu \in V(C^u), e \in \Ecirc(u)$, indicating the degree other than that of the ring edge $e$ at $\mu$ in $C^u$;
\item integer variables $\dg_{[e], [\mu], -}^{\re, u}, u \in \Vcirc, \mu \in V(C^u), e \in \Ecirc(u)$, indicating the augmented degree for $\mu$ because of the ring edge $e$;
\end{itemize}

\noindent
{\bf Constraints.}

\noindent
For each $u \in \Vcirc, \mu \in V(C^u)$:
\begin{align}
&0 \leq \dg_{[e], [\mu], +}^{\re, u} \le 4 \sum_{\nu \in (E(C^u))(\mu)} x_{[e], [\nu]}^{\re, u} && e \in \Ecirc(u); \nonumber \\
&0 \leq \dg_{[e], [\mu], -}^{\re, u} \le 4 \sum_{\nu \in (E(C^u))(\mu)} x_{[e], [\nu]}^{\re, u} && e \in \Ecirc(u); \\
& 4(x_{[e],[\nu]}^{\re, u} - 1) \le \dg_{[e],[\mu], +}^{\re, u} -1 - \sum_{e' \in \barEcirc(u)} x_{[e'], [\mu]}^{\rn, u} \le 4(1 - x_{[e],[\nu]}^{\re, u}), && e \in \Ecirc(u), \nu \in (E(C^u))(\mu);
\end{align}

\noindent
For each $e=uu'\in\Ecirc$ such that $[u]<[u']$:
\begin{align}
 3 (x_{[e],[\nu]}^{\re, u} + x_{[e],[\nu']}^{\re, u'} - 2) \le \dg_{[e], i_1, -}^{\re, u} - \dg_{[e],j_2, +}^{\re, u'}  \le 3 (2-x_{[e],[\nu]}^{\re, u} - x_{[e],[\nu']}^{\re, u'}), & \nonumber \\
  3 (x_{[e],[\nu]}^{\re, u} + x_{[e],[\nu']}^{\re, u'} - 2) \le \dg_{[e], i_2, -}^{\re, u} - \dg_{[e],j_1, +}^{\re, u'}  \le 3 (2-x_{[e],[\nu]}^{\re, u} - x_{[e],[\nu']}^{\re, u'}), & \nonumber \\
   3 (x_{[e],[\nu]}^{\re, u} + x_{[e],[\nu']}^{\re, u'} - 2) \le \dg_{[e], j_2, -}^{\re, u'} - \dg_{[e],i_1, +}^{\re, u}  \le 3 (2-x_{[e],[\nu]}^{\re, u} - x_{[e],[\nu']}^{\re, u'}), & \nonumber \\
    3 (x_{[e],[\nu]}^{\re, u} + x_{[e],[\nu']}^{\re, u'} - 2) \le \dg_{[e], j_1, -}^{\re, u'} - \dg_{[e],i_2, +}^{\re, u}  \le 3 (2-x_{[e],[\nu]}^{\re, u} - x_{[e],[\nu']}^{\re, u'}), & \nonumber \\
   \nu=u_{i_1}u_{i_2}\in E(C^u),\ \nu'=u'_{j_1}u'_{j_2}\in E(C^{u'})&\nonumber\\ 
   \textrm{such\ that\ }i_1<i_2,\ j_1<j_2; &
\end{align}

\noindent
For each $u \in \Vcirc, \mu \in V(C^u)$:
\begin{align}
 \sum_{d \in [1,4]} \ddg(u, [\mu]; d) &  = \sum_{\xi \in \Xi^u, |\xi| \ge [\mu]} x_{[\xi]}^u;  \\
 2(\sum_{\xi \in \Xi^u, |\xi| \ge [\mu]} x_{[\xi]}^u - 1) & \leq  \sum_{d \in [1,4]} d \cdot \ddg(u, [\mu]; d) - \nonumber \\
 & (2 + \sum_{\psi \in \fr^u} \dgnoH(\psi) \dfr(u, [\mu]; [\psi])  + \sum_{e' \in \barEcirc(u)} x_{[e'], [\mu]}^{\rn, u} +\sum_{e\in\Ecirc(u)}  
 \dg_{[e], [\mu], -}^{\re, u}); \nonumber \\
 \sum_{d \in [1,4]} d \cdot \ddg(u, [\mu]; d) & \leq \ 2 + \sum_{\psi \in \fr^u} \dgnoH(\psi) \dfr(u, [\mu]; [\psi]) \nonumber \\
& + \sum_{e' \in \barEcirc(u)} x_{[e'], [\mu]}^{\rn, u} +\sum_{e\in\Ecirc(u)} \dg_{[e], [\mu], -}^{\re, u}); \\
 \sum_{d \in [1,4]} \ddgint(u, [\mu]; d) & = \sum_{\xi \in \Xi^u, |\xi| \ge [\mu]} x_{[\xi]}^u;  \\
 2(\sum_{\xi \in \Xi^u, |\xi| \ge [\mu]} x_{[\xi]}^u - 1) & \leq
 \sum_{d \in [1,4]} d \cdot \ddgint(u, [\mu]; d) -  \ (2 + \sum_{e' \in \barEcirc(u)} x_{[e'], [\mu]}^{\rn, u} + \sum_{e\in\Ecirc(u)} \dg_{[e], [\mu], -}^{\re, u}); \nonumber \\
 \sum_{d \in [1,4]} d \cdot \ddgint(u, [\mu]; d) & \leq  \ 2 + \sum_{e' \in \barEcirc(u)} x_{[e'], [\mu]}^{\rn, u} + \sum_{e\in\Ecirc(u)} \dg_{[e], [\mu], -}^{\re, u});
\end{align}

\noindent
For each $v \in V(T) \setminus \Vcirc$:
\begin{align}
& \sum_{d \in [1,4]} \ddg(v; d)  =  1;  \\
& \sum_{d \in [1,4]} d \cdot \ddg(v; d) = |N_T(v)| + \sum_{\psi \in \fr^v} \dgnoH(\psi) \dfr(v; [\psi]); \\
& \sum_{d \in [1,4]} \ddgint(v; d)  =  1;  \\
& \sum_{d \in [1,4]} d \cdot \ddgint(v; d) = |N_T(v)|;
\end{align}

\noindent
For each $e=uu'\in\Ecirc$ such that $[u]<[u']$:
\begin{align}
\dg([e]; d) - \ddg(u, i_1; d) - \ddg(u, i_2; d) \le 2(1 - x_{[e], [\nu]}^{\re, u}), & \nonumber \\
\dg([e]; d) - \ddg(u, i_1; d) - \ddg(u, i_2; d) \ge 2(x_{[e], [\nu]}^{\re, u} - 1), &\nu = u_{i_1}u_{i_2}\in E(C^u), d \in [1,4]  \\
\dgint([e]; d) - \ddgint(u, i_1; d) - \ddgint(u, i_2; d) \le 2(1-  x_{[e], [\nu]}^{\re, u}), & \nonumber \\
\dgint([e]; d) - \ddgint(u, i_1; d) - \ddgint(u, i_2; d) \ge 2( x_{[e], [\nu]}^{\re, u} - 1), & \nu = u_{i_1}u_{i_2}\in E(C^u), d \in [1,4] 
\end{align}

\noindent
For each $d \in [1,4]$:
\begin{align}
& \dg(d) = \sum_{u \in \Vcirc} \sum_{\mu \in V(C^u)} \ddg(u, [\mu]; d) + 
\sum_{v \in V(T) \setminus \Vcirc} \ddg(v; d) - \sum_{e \in \Ecirc} \dg([e];d); \\
& \dgint(d) = \sum_{u \in \Vcirc} \sum_{\mu \in V(C^u)} \ddgint(u, [\mu]; d) + 
\sum_{v \in V(T) \setminus \Vcirc} \ddgint(v; d) - \sum_{e \in \Ecirc} \dgint([e];d); 
\end{align}

\subsection{Assigning Bond-Multiplicity}

\noindent
{\bf Variables.}
\begin{itemize}
\item Integer variables $\beta_i^u \in [0,3], u \in \Vcirc, i \in [1, 2\Cmax-\Cmin]$, that stores the bond-multiplicty of the edge $e_i$ in $C^u$;
\item integer variables $\beta_{[e]} \in [1,3], e \in E(T)$, that stores the bond-multiplicty of the edge $e$;
\item binary variables $\dbeta(u, i; m), u \in \Vcirc, i \in [1, 2\Cmax-\Cmin], m \in [1,3]$, $\dbeta(u,i;m)=1 \Leftrightarrow \beta_i^u = m$;
\item binary variables $\dbeta([e]; m), e \in E(T), m \in [1,3]$, $\dbeta([e];m)=1 \Leftrightarrow \beta_{[e]} = m$;
\item integer variables $\bd(m), m \in [1,3]$, that stores the number of edges with bond-multiplicty $m$;
\end{itemize}

\noindent
{\bf Constraints.}

\noindent

\begin{align}
& e_i^u \le \beta_i^u \le 3 e_i^u, && u \in \Vcirc, i \in [1, 2\Cmax-\Cmin]; \\
& 1 \le \beta_{[e]} \le 3, && e \in E(T);
\end{align}

\begin{align}
& \sum_{m \in [1,3]} \dbeta(u, i;m) = e_i^u, & \sum_{m \in [1,3]} m \cdot \dbeta(u,i;m) = \beta_i^u, && u \in \Vcirc, i \in [1, 2\Cmax-\Cmin]; \\
& \sum_{m \in [1,3]} \dbeta([e]; m) = 1, & \sum_{m \in [1,3]} m \cdot \dbeta([e]; m) = \beta_{[e]}, && e \in E(T);
\end{align}

\noindent
For each $u \in \Vcirc, e \in \Ecirc(u)$:
\begin{align}
& 3(x_{[e],i}^{\re, u} - 1) \le \beta_i^u - \beta_{[e]} \le 3(1 - x_{[e],i}^{\re, u}), && i \in [1, 2\Cmax-\Cmin];
\end{align}

\noindent
For each $m \in [1,3]$:
\begin{align}
& \bd(m) = \sum_{u \in \Vcirc} \sum_{i \in [1, 2\Cmax-\Cmin]} \dbeta(u,i;m) + \sum_{e' \in E(T) \setminus \Ecirc} \dbeta([e'];m) - \sum_{e \in \Ecirc} \dbeta([e];m); &&
\end{align}

\subsection{Assigning Chemical Elements and Valence Condition}
{\bf Constants.}
\begin{itemize}
\item A set $\Lambda$ consisting of all available chemical elements;
\item functions $\alphar(\psi), \valexfr(\psi), \eledegfr(\psi), \nta(\psi)$ denoting the chemical element of the root, root valence, ion-valence, number of non-root chemical element $\tta$ of the fringe-tree $\psi$, respectively;
\item functions $\val(\tta), \mass^*(\tta)$ denoting the valence and mass of the chemical element $\tta$, respectively;
\item integers $\nainLB([\tta]), \nainUB([\tta]) \in [0, \nUB], \tta \in \Lambda$, that represent the lower and upper bounds of the chemical element $\tta$ in the interior part, respectively;
\item integers $\naexLB([\tta]), \naexUB([\tta]) \in [0, \nUB], \tta \in \Lambda$, that represent the lower and upper bounds of the chemical element $\tta$ in the exterior part, respectively; 
\item integers $\naLB([\tta]), \naUB([\tta]) \in [0, \nUB], \tta \in \Lambda$, that represent the lower and upper bounds of the chemical element $\tta$ in $G_\MT$, respectively;
\item a positive constant $M_{\ms} \in \mathbb{R}_+$ that represents a sufficiently large number;
\end{itemize}

\noindent
{\bf Variables.}
\begin{itemize}
\item Integer variables $\alpha(u, [\mu]), u \in \Vcirc, \mu \in V(C^u)$, that represents the chemical element assigned to the vertex $\mu$ in $C^u$;
\item integer variables $\alpha(v), v \in V(T) \setminus \Vcirc$, that represents the chemical element assigned to the vertex $v$;
\item binary variables $\dalpha(u, [\mu]; [\tta]), u \in \Vcirc, \mu \in V(C^u), \tta \in \Lambda$, $\dalpha(u,[\mu]; [\tta]) = 1 \Leftrightarrow \alpha(u, [\mu]) = [\tta]$;
\item binary variables $\dalpha(v; [\tta]), v \in V(T) \setminus \Vcirc, \tta \in \Lambda$, $\dalpha(v; [\tta]) = 1 \Leftrightarrow \alpha(v) = [\tta]$;
\item integer variables $\beta_{[e'], [\mu]}^{\rn, u}, u \in \Vcirc, \mu \in V(C^u), e' \in \barEcirc(u)$, indicating the bond-multiplicity assigned to the non-ring edge $e'$ at vertex $\mu$;
\item integer variables $\beta_{[e], [\mu], +}^{\re, u}, u \in \Vcirc, \mu \in V(C^u), e \in \Ecirc(u)$, indicating the bond-multiplicity other than that of the ring edge $e$ at $\mu$ in $C^u$;
\item integer variables $\beta_{[e], [\mu], -}^{\re, u}, u \in \Vcirc, \mu \in V(C^u), e \in \Ecirc(u)$, indicating the augmented bond-multiplicity for $\mu$ because of the ring edge $e$;
\item integer variables $\na([e]; [\tta])\in[0,2], e \in \Ecirc, \tta \in \Lambda$, that stores the number of chemical element $\tta$ used in the ring edge $e$;
\item integer variables $\nain([\tta]) \in [\nainLB([\tta]), \nainUB([\tta])], \tta \in \Lambda$, that stores the number of chemical element $\tta$ in the interior part;
\item integer variables $\naex([\tta]) \in [\naexLB([\tta]), \naexUB([\tta])], \tta \in \Lambda$, that stores the number of chemical element $\tta$ in the exterior part;
\item integer variables $\na([\tta]) \in [\naLB([\tta]), \naUB([\tta])], \tta \in \Lambda$, that stores the number of chemical element $\tta$ in $G_\MT$;
\item binary variables $\datom(i), i \in [\nLB+\naLB([\ttH]), \nUB+\naUB([\ttH])]$, $\datom(i) = 1 \Leftrightarrow n_G = i$;
\item integer variable $\Mass$ that represents the total mass of $G_\MT$;
\item real variable $\overline{\ms}$ that represents the average mass of $G_\MT$;
\end{itemize}

\noindent
{\bf Constraints.}

\noindent
For each $u \in \Vcirc, \mu \in V(C^u)$:
\begin{align}
& \alpha(u, [\mu]) = \sum_{\psi \in \fr^u} [\alphar(\psi)] \cdot \dfr(u, [\mu]; [\psi]); && \\
& \sum_{\tta \in \Lambda} \dalpha(u, [\mu]; [\tta]) = \sum_{\xi \in \Xi^u, |\xi| \ge [\mu]} x_{[\xi]}^u; && \\
& \sum_{\tta \in \Lambda} [\tta] \cdot \dalpha(u, [\mu]; [\tta]) = \alpha(u,[\mu]); && 
\end{align}

\noindent
For each $v \in V(T) \setminus \Vcirc$:
\begin{align}
& \alpha(v) = \sum_{\psi \in \fr^v} [\alphar(\psi)] \cdot \dfr(v; [\psi]); && \\
& \sum_{\tta \in \Lambda} \dalpha(v; [\tta]) = 1; && \\
& \sum_{\tta \in \Lambda} [\tta] \cdot \dalpha(v; [\tta]) = \alpha(v); && 
\end{align}

\noindent
For each $u \in \Vcirc, \mu \in V(C^u)$:
\begin{align}
& 0 \leq \beta_{[e'], [\mu]}^{\rn, u} \le 3 x_{[e'], [\mu]}^{\rn, u}, && e' \in \barEcirc(u); \\
& 3(x_{[e'],[\mu]}^{\rn, u} - 1) \le \beta_{[e']} - \beta_{[e'], [\mu]}^{\rn, u} \le 3(1 - x_{[e'],[\mu]}^{\rn, u}), && e' \in \barEcirc(u);\\
&0 \leq \beta_{[e], [\mu], +}^{\re, u} \le 6 \sum_{\nu \in (E(C^u))(\mu)} x_{[e], [\nu]}^{\re, u} && e \in \Ecirc(u); \nonumber \\
&0 \leq \beta_{[e], [\mu], -}^{\re, u} \le 6 \sum_{\nu \in (E(C^u))(\mu)} x_{[e], [\nu]}^{\re, u} && e \in \Ecirc(u); \\
& 6(x_{[e],[\nu]}^{\re, u} - 1) \le \beta_{[e],[\mu], +}^{\re, u} + \beta_{[e]} - \sum_{\nu' \in (E(C^u))(\mu)} \beta_{[\nu']}^u  \nonumber \\
& \ \ \ \ \ \  \ \ \  \ \ \ \  \ \ \ \ \ \ \ \ \ - \sum_{e' \in \barEcirc(u)} \beta_{[e'], [\mu]}^{\rn, u} \le 6(1 - x_{[e],[\nu]}^{\re, u}),  && e \in \Ecirc(u), \nu \in (E(C^u))(\mu); 
\end{align}

\noindent
For each $e=uu'\in\Ecirc$ such that $[u]<[u']$:
\begin{align}
 3 (x_{[e],[\nu]}^{\re, u} + x_{[e],[\nu']}^{\re, u'} - 2) \le \beta_{[e], i_1, -}^{\re, u} - \beta_{[e],j_2, +}^{\re, u'}  \le 3 (2-x_{[e],[\nu]}^{\re, u} - x_{[e],[\nu']}^{\re, u'}), & \nonumber \\
  3 (x_{[e],[\nu]}^{\re, u} + x_{[e],[\nu']}^{\re, u'} - 2) \le \beta_{[e], i_2, -}^{\re, u} - \beta_{[e],j_1, +}^{\re, u'}  \le 3 (2-x_{[e],[\nu]}^{\re, u} - x_{[e],[\nu']}^{\re, u'}), & \nonumber \\
   3 (x_{[e],[\nu]}^{\re, u} + x_{[e],[\nu']}^{\re, u'} - 2) \le \beta_{[e], j_2, -}^{\re, u'} - \beta_{[e],i_1, +}^{\re, u}  \le 3 (2-x_{[e],[\nu]}^{\re, u} - x_{[e],[\nu']}^{\re, u'}), & \nonumber \\
    3 (x_{[e],[\nu]}^{\re, u} + x_{[e],[\nu']}^{\re, u'} - 2) \le \beta_{[e], j_1, -}^{\re, u'} - \beta_{[e],i_2, +}^{\re, u}  \le 3 (2-x_{[e],[\nu]}^{\re, u} - x_{[e],[\nu']}^{\re, u'}), & \nonumber \\
   \nu=u_{i_1}u_{i_2}\in E(C^u),\ \nu'=u'_{j_1}u'_{j_2}\in E(C^{u'})&\nonumber\\ 
   \textrm{such\ that\ }i_1<i_2,\ j_1<j_2; &
\end{align}

\begin{align}
\sum_{\tta \in \Lambda} \val(\tta) \cdot \dalpha(u, [\mu];[\tta]) & = \sum_{\nu \in (E(C^u))(\mu)} \beta_{[\nu]}^u + \sum_{e' \in \barEcirc(u)} \beta_{[e'], [\mu]}^{\rn, u} + \sum_{e \in \Ecirc(u)} \beta_{[e],[\mu],-}^{\re, u} \nonumber \\
& + \sum_{\psi \in \fr^u} (\valexfr(\psi) - \eledegfr(\psi)) \dfr(u, [\mu];[\psi]), & u \in \Vcirc, \mu \in V(C^u); \\
\sum_{\tta \in \Lambda} \val(\tta) \cdot \dalpha(v;[\tta]) & = \sum_{e' \in \barEcirc(v)} \beta_{[e']}
+ \sum_{\psi \in \fr^v} (\valexfr(\psi) - \eledegfr(\psi)) \dfr(v;[\psi]), & v \in V(T) \setminus \Vcirc; 
\end{align}

\noindent
For each $\tta \in \Lambda$:
\begin{align}
& \na([e];[\tta]) = \sum_{\psi \in \fr^\MT, \alphar(\psi) = \tta} \fc([e]; [\psi]), & e \in \Ecirc; \\
& \nain([\tta]) = \sum_{u \in \Vcirc} \sum_{\mu \in V(C^u)} \dalpha(u, [\mu];[\tta]) + \sum_{v \in V(T) \setminus \Vcirc} \dalpha(v;[\tta]) - \sum_{e \in \Ecirc} \na([e]; [\tta]); \\
& \naex([\tta]) = \sum_{\psi \in \fr^\MT} \nta(\psi) \cdot \fc([\psi]); \\
& \na([\tta]) = \nain([\tta]) + \naex([\tta]);
\end{align}

\begin{align}
\Mass = \sum_{\tta \in \Lambda} \mass^*(\tta) \cdot \na([\tta]); & \\
\sum_{i \in [\nLB+ \naLB([\ttH]), \nUB+\naUB([\ttH])]} \datom(i) = 1; & \\
\sum_{i \in [\nLB+ \naLB([\ttH]), \nUB+\naUB([\ttH])]} i \cdot \datom(i) = n_G + \naex([\ttH]); & \\
M_\ms(\datom(i) - 1) \le \overline{\ms} - \frac{\Mass}{ i} \le M_\ms(1 - \datom(i)), & i \in [\nLB+ \naLB([\ttH]), \nUB+\naUB([\ttH])];
\end{align}

\subsection{Descriptors for the Number of Adjacency-configurations}
{\bf Constants.}
\begin{itemize}
\item A set $\Gacin$ consisting of available adjacency-configurations;
\item integers $\acLB(\gamma), \acUB(\gamma) \in [0, \nUB + |\Vcirc| - 1], \gamma \in \Gacin$, that represent the lower and upper bounds of the adjacency-configuration $\gamma$ in $G_\MT$, respectively;
\end{itemize}

\noindent
Here, for an adjacency-configuration $\gamma = (\tta, \ttb, m)$, we denote $\ogamma := (\ttb, \tta, m)$. The set $\Gacin$ is supposed to satisfy $\gamma \in \Gacin \Rightarrow \ogamma \in \Gacin$.

\noindent
{\bf Variables.}
\begin{itemize}
\item Binary variables $\dac(u, [\nu]; [\gamma]), u \in \Vcirc, \nu \in E(C^u), \gamma \in \Gacin$, indicating whether the edge $\nu$ has adjacency-configuration $\gamma$;
\item binary variables $\dac([e]; [\gamma]), e \in E(T), \gamma \in \Gacin$, indicating whether the edge $e$ has adjacency-configuration $\gamma$;
\item integer variables $\acin([\gamma]) \in [\acLB(\gamma), \acUB(\gamma)], \gamma \in \Gacin$, that stores the adjacency-configurations;
\end{itemize}

\noindent
{\bf Constraints.}

\noindent
For each $u \in \Vcirc, \nu = u_iu_j \in E(C^u)$ such that $i < j$:
\begin{align}
& \sum_{\gamma \in \Gacin} \dac(u, [\nu]; [\gamma]) = e_{[\nu]}^u; & \\
& \sum_{\gamma=(\tta,\ttb,m) \in \Gacin} m \cdot \dac(u, [\nu]; [\gamma]) - \beta_{[\nu]}^u \geq 3(e_{[\nu]}^u - 1) ; & \nonumber \\
& \sum_{\gamma=(\tta,\ttb,m) \in \Gacin} m \cdot \dac(u, [\nu]; [\gamma]) \leq \beta_{[\nu]}^u ; & \\
& \sum_{\gamma=(\tta,\ttb,m) \in \Gacin} [\tta] \cdot \dac(u, [\nu]; [\gamma]) - \alpha(u,i) \geq |\Lambda| (e_{[\nu]}^u - 1); & \nonumber \\
& \sum_{\gamma=(\tta,\ttb,m) \in \Gacin} [\tta] \cdot \dac(u, [\nu]; [\gamma]) \leq \alpha(u,i); & \\
& \sum_{\gamma=(\tta,\ttb,m) \in \Gacin} [\ttb] \cdot \dac(u, [\nu]; [\gamma]) - \alpha(u,j) \geq |\Lambda| (e_{[\nu]}^u - 1); & \nonumber \\
& \sum_{\gamma=(\tta,\ttb,m) \in \Gacin} [\ttb] \cdot \dac(u, [\nu]; [\gamma]) \leq \alpha(u,j); & 
\end{align}

\noindent
For each $e \in E(T)$:
\begin{align}
& \sum_{\gamma \in \Gacin} \dac([e]; [\gamma]) = 1; & \\
& \sum_{\gamma=(\tta,\ttb,m) \in \Gacin} m \cdot \dac([e]; [\gamma]) = \beta_{[e]}; &
\end{align}

\noindent
For each non-ring edge $e' =uv \in E(T) \setminus \Ecirc$ such that $[u] < [v]$:
\begin{align}
& \sum_{\gamma = (\tta, \ttb, m) \in \Gacin} [\tta] \cdot \dac([e']; [\gamma]) = \alpha(u), & & \textrm{if\ } u \notin \Vcirc; \nonumber \\
& |\Lambda| (x_{[e'], [\mu]}^{\rn, u} - 1) \le \sum_{\gamma = (\tta, \ttb, m) \in \Gacin} [\tta] \cdot \dac([e'];[\gamma]) - \alpha(u,[\mu]) \le |\Lambda| (1 - x_{[e'], [\mu]}^{\rn, u} ), & \mu \in V(C^u), \ & \textrm{if\ } u \in \Vcirc; \\
& \sum_{\gamma = (\tta, \ttb, m) \in \Gacin} [\ttb] \cdot \dac([e']; [\gamma]) = \alpha(v), & & \textrm{if\ } v \notin \Vcirc; \nonumber \\
& |\Lambda| (x_{[e'], [\mu]}^{\rn, v} - 1) \le \sum_{\gamma = (\tta, \ttb, m) \in \Gacin} [\ttb] \cdot \dac([e'];[\gamma]) - \alpha(v,[\mu]) \le |\Lambda| (1 - x_{[e'], [\mu]}^{\rn, v} ), & \mu \in V(C^v), \ & \textrm{if\ } v \in \Vcirc; 
\end{align}

\noindent
For each $e=uu'\in\Ecirc$ such that $[u]<[u']$:
\begin{align}
  |\Lambda| (x_{[e], [\nu]}^{\re, u} - 1) \le   \sum_{\gamma=(\tta, \ttb, m) \in \Gacin} [\tta] \cdot \dac([e];[\gamma]) - \alpha(u, i_1) \le  & |\Lambda| (1 - x_{[e], [\nu]}^{\re, u}),   \nonumber \\
  |\Lambda| (x_{[e], [\nu]}^{\re, u} - 1) \le   \sum_{\gamma=(\tta, \ttb, m) \in \Gacin} [\ttb] \cdot \dac([e];[\gamma]) - \alpha(u, i_2) \le & |\Lambda| (1 - x_{[e], [\nu]}^{\re, u}),   \nonumber \\ 
 &  \nu = u_{i_1}u_{i_2} \in E(C^u) \ \textrm{such\ that\ } i_1 < i_2; 
\end{align}

\noindent
For each $\gamma \in \Gacin$:
\begin{align}
\acin([\gamma]) = & \sum_{u \in \Vcirc} \sum_{\nu \in E(C^u)} (\dac(u, [\nu];[\gamma]) + \dac(u, [\nu]; [\ogamma] ) \nonumber \\
& + \sum_{e' \in E(T) \setminus \Ecirc} (\dac([e'];[\gamma])+ \dac([e'];[\ogamma])) \nonumber \\
& - \sum_{e \in \Ecirc} (\dac([e];[\gamma]) + \dac([e];[\ogamma])), & \textrm{if\ } \gamma \neq \ogamma; \nonumber \\
\acin([\gamma]) = & \sum_{u \in \Vcirc} \sum_{\nu \in E(C^u)} \dac(u, [\nu];[\gamma])  + \sum_{e' \in E(T) \setminus \Ecirc} \dac([e'];[\gamma]) \nonumber \\
& - \sum_{e \in \Ecirc} \dac([e];[\gamma]) , & \textrm{if\ } \gamma = \ogamma; 
\end{align}

\subsection{Descriptors for the Number of Edge-configurations}
{\bf Constants.}
\begin{itemize}
\item A set $\Gecin$ consisting of available edge-configurations;
\item integers $\ecLB(\tau), \ecUB(\tau) \in [0, \nUB + |\Vcirc| - 1], \gamma \in \Gecin$, that represent the lower and upper bounds of the adjacency-configuration $\tau$ in $G_\MT$, respectively;
\end{itemize}

\noindent
Here, for an edge-configuration $\tau = (\tta d, \ttb d', m)$, we denote $\otau := (\ttb d', \tta d, m)$. The set $\Gecin$ is supposed to satisfy $\tau \in \Gecin \Rightarrow \otau \in \Gecin$.

\noindent
{\bf Variables.}
\begin{itemize}
\item Binary variables $\dec(u, [\nu]; [\tau]), u \in \Vcirc, \nu \in E(C^u), \tau \in \Gecin$, indicating whether the edge $\nu$ has edge-configuration $\tau$;
\item binary variables $\dec([e]; [\tau]), e \in E(T), \tau \in \Gecin$, indicating whether the edge $e$ has edge-configuration $\tau$;;
\item integer variables $\ecin([\tau]) \in [\ecLB(\tau), \ecUB(\tau)], \tau \in \Gecin$, that stores the edge-configurations;
\end{itemize}

\noindent
{\bf Constraints.}

\noindent
For each $u \in \Vcirc, \nu = u_iu_j \in E(C^u)$ such that $i < j$:
\begin{align}
& \sum_{\tau \in \Gecin} \dec(u, [\nu]; [\tau]) = e_{[\nu]}^u; & \\
& \sum_{\tau = (\tta d, \ttb d', m) \in \Gecin} [(\tta, \ttb, m)] \cdot \dec(u, [\nu]; [\tau]) = \sum_{\gamma \in \Gacin} [\gamma] \cdot \dac(u, [\nu];[\gamma]); & \\
& \sum_{\tau=(\tta d,\ttb d',m) \in \Gecin} d \cdot \dec(u, [\nu]; [\tau]) - \sum_{d \in [1,4]} d \cdot \ddg(u, i; d) \geq 4(e_{[\nu]}^u - 1); \nonumber \\
& \sum_{\tau=(\tta d,\ttb d',m) \in \Gecin} d \cdot \dec(u, [\nu]; [\tau]) \leq \sum_{d \in [1,4]} d \cdot \ddg(u, i; d); & \\
& \sum_{\tau=(\tta d,\ttb d',m) \in \Gecin} d' \cdot \dec(u, [\nu]; [\tau]) - \sum_{d' \in [1,4]} d' \cdot \ddg(u, j; d') \geq 4(e_{[\nu]}^u - 1); \nonumber \\
& \sum_{\tau=(\tta d,\ttb d',m) \in \Gecin} d' \cdot \dec(u, [\nu]; [\tau]) \leq \sum_{d' \in [1,4]} d' \cdot \ddg(u, j; d'); & 
\end{align}

\noindent
For each $e \in E(T)$:
\begin{align}
& \sum_{\tau \in \Gecin} \dec([e];[\tau]) = 1; & \\
& \sum_{\tau = (\tta d, \ttb d', m) \in \Gecin} [(\tta, \ttb, m)] \cdot \dec([e]; [\tau]) = \sum_{\gamma \in \Gacin} [\gamma] \cdot \dac([e]; [\gamma]); & 
\end{align}

\noindent
For each non-ring edge $e' =uv \in E(T) \setminus \Ecirc$ such that $[u] < [v]$:
\begin{align}
& \sum_{\tau = (\tta d, \ttb d', m) \in \Gecin} d \cdot \dec([e']; [\tau]) = \sum_{d \in [1,4]} d \cdot \ddg(u;d), & & \textrm{if\ } u \notin \Vcirc; \nonumber \\
& \sum_{\tau = (\tta d, \ttb d', m) \in \Gecin} d \cdot \dec([e'];[\tau]) - \sum_{d \in [1,4]} d \cdot \ddg(u, [\mu]; d) \ge 4 (x_{[e'], [\mu]}^{\rn, u} -1), & \nonumber \\
& \sum_{\tau = (\tta d, \ttb d', m) \in \Gecin} d \cdot \dec([e'];[\tau]) - \sum_{d \in [1,4]} d \cdot \ddg(u, [\mu]; d) \le 4 (1 - x_{[e'], [\mu]}^{\rn, u} ), & \mu \in V(C^u), \ & \textrm{if\ } u \in \Vcirc; \\
& \sum_{\tau = (\tta d, \ttb d', m) \in \Gecin} d' \cdot \dac([e']; [\tau]) = \sum_{d' \in [1,4]} d' \cdot \ddg(v;d), & & \textrm{if\ } v \notin \Vcirc; \nonumber \\
& \sum_{\tau = (\tta d, \ttb d', m) \in \Gecin} d' \cdot \dec([e'];[\tau]) - \sum_{d' \in [1,4]} d' \cdot \ddg(v, [\mu]; d') \ge 4 (x_{[e'], [\mu]}^{\rn, v} -1), & \nonumber \\
& \sum_{\tau = (\tta d, \ttb d', m) \in \Gecin} d' \cdot \dec([e'];[\tau]) - \sum_{d' \in [1,4]} d' \cdot \ddg(v, [\mu]; d') \le 4 (1 - x_{[e'], [\mu]}^{\rn, v} ), & \mu \in V(C^v), \  & \textrm{if\ } v \in \Vcirc; 
\end{align}

\noindent
For each $e=uu'\in\Ecirc$ such that $[u]<[u']$:
\begin{align}
4 (x_{[e], [\nu]}^{\re, u} - 1) \le   \sum_{\tau=(\tta d, \ttb d', m) \in \Gecin} d \cdot \dec([e];[\tau]) & - \sum_{d \in [1,4]} d \cdot \ddg(u, i_1; d) \le   4(1 - x_{[e], [\nu]}^{\re, u}),   \nonumber \\
4 (x_{[e], [\nu]}^{\re, u} - 1) \le   \sum_{\tau=(\tta d, \ttb d', m) \in \Gecin} d' \cdot \dec([e];[\tau]) & - \sum_{d' \in [1,4]} d' \cdot \ddg(u, i_2; d) \le  4 (1 - x_{[e], [\nu]}^{\re, u}),   \nonumber \\ 
 &  \nu = u_{i_1}u_{i_2} \in E(C^u) \ \textrm{such\ that\ } i_1 < i_2; 
\end{align}

\noindent
For each $\tau \in \Gecin$:
\begin{align}
\ecin([\tau]) = & \sum_{u \in \Vcirc} \sum_{\mu \in V(C^u)} (\dec(u, [\mu];[\tau]) + \dec(u, [\mu]; [\otau] ) \nonumber \\
& + \sum_{e' \in E(T) \setminus \Ecirc} (\dec([e'];[\tau])+ \dec([e'];[\otau])) \nonumber \\
& - \sum_{e \in \Ecirc} (\dec([e];[\tau]) + \dec([e];[\otau])), & \textrm{if\ } \tau \neq \otau; \nonumber \\
\ecin([\tau]) = & \sum_{u \in \Vcirc} \sum_{\mu \in V(C^u)} \dec(u, [\mu];[\tau])  + \sum_{e' \in E(T) \setminus \Ecirc} \dec([e'];[\tau]) \nonumber \\
& - \sum_{e \in \Ecirc} \dec([e];[\tau]) , & \textrm{if\ } \tau = \otau; 
\end{align}

\end{document}